\documentclass{article}
\usepackage[utf8]{inputenc}

\usepackage{pgfplots}       

\pgfplotsset{compat=1.18}

\usepackage[pdfencoding=auto]{hyperref}

\usepackage{bm}

\usepackage{geometry}
\usepackage{graphicx}

\usepackage[citestyle=authoryear, backend=bibtex, sorting=nyt, natbib=true, url=false, doi=false]{biblatex}
\usepackage{amsmath}
\usepackage{amssymb} 
\usepackage{nicefrac}  
\usepackage{float} 
\usepackage{amsthm} 
\newtheorem{proposition}{Proposition}[section]
\newtheorem{definition}{Definition}[section]

\usepackage{hyperref} 

\usepackage{xcolor} 
\usepackage{threeparttable}
\usepackage{thmtools} 

\usepackage{cleveref} 
\crefname{hyp}{hypothesis}{hypotheses} 
\Crefname{hyp}{Hypothesis}{Hypotheses} 

\usepackage{ragged2e} 

\usepackage{booktabs} 
\usepackage{multirow} 

\usepackage{algorithm}
\usepackage{algpseudocode}

\algnewcommand{\IfThen}[2]{
	\State \algorithmicif\ #1\ \algorithmicthen\ #2}
\algnewcommand{\LineComment}[1]{\State \(\triangleright\) #1}

\algnewcommand{\InlineFor}[2]{
	\State \algorithmicfor\ #1\ \algorithmicendfor\ #2}

 \makeatletter
\AfterEndEnvironment{algorithm}{\let\@algcomment\relax}
\AtEndEnvironment{algorithm}{\kern2pt\hrule\relax\vskip3pt\@algcomment}
\let\@algcomment\relax
\newcommand\algcomment[1]{\def\@algcomment{\footnotesize#1}}

\renewcommand\fs@ruled{\def\@fs@cfont{\bfseries}\let\@fs@capt\floatc@ruled
  \def\@fs@pre{\hrule height.8pt depth0pt \kern2pt}%
  \def\@fs@post{}%
  \def\@fs@mid{\kern2pt\hrule\kern2pt}%
  \let\@fs@iftopcapt\iftrue}
\makeatother

\usepackage{booktabs}

\usepackage[parfill]{parskip} 

\usepackage{authblk}

\usepackage{algorithm}
\usepackage{algpseudocode}

\providecommand{\keywords}[1]
{
  \small	
  \textbf{\textit{Keywords---}} #1
}

\title{Study of Robust Features in Formulating Guidance for Heuristic Algorithms for Solving the Vehicle Routing Problem}

\author[1]{Bachtiar Herdianto}
\author[1]{Romain Billot}
\author[2]{Flavien Lucas}
\author[3]{Marc Sevaux}

\date{\color{blue} 
      \nolinkurl{bachtiar.herdianto@imt-atlantique.fr}$^{*1}$\\
      \nolinkurl{romain.billot@imt-atlantique.fr}$^1$\\
      \nolinkurl{flavien.lucas@imt-nord-europe.fr}$^2$\\
      \nolinkurl{marc.sevaux@univ-ubs.fr}$^3$}

\affil[1]{IMT Atlantique, Lab-STICC (UMR 6285, CNRS), Brest, France}
\affil[2]{IMT Nord Europe, CERI Systèmes Numériques, Douai, France}
\affil[3]{Université Bretagne Sud, Lab-STICC (UMR 6285, CNRS), Lorient, France}

\begin{document}

\maketitle

\begin{abstract}
The Vehicle Routing Problem (VRP) is a complex optimization problem with numerous real-world applications, mostly solved using metaheuristic algorithms due to its $\mathcal{NP}$-Hard nature. Traditionally, these metaheuristics rely on human-crafted designs developed through empirical studies. However, recent research shows that machine learning methods can be used the structural characteristics of solutions in combinatorial optimization, thereby aiding in designing more efficient algorithms, particularly for solving VRP. Building on this advancement, this study extends the previous research by conducting a sensitivity analysis using multiple classifier models that are capable of predicting the quality of VRP solutions. Hence, by leveraging explainable AI, this research is able to extend the understanding of how these models make decisions. Finally, our findings indicate that while feature importance varies, certain features consistently emerge as strong predictors. Furthermore, we propose a unified framework able of ranking feature impact across different scenarios to illustrate this finding. These insights highlight the potential of feature importance analysis as a foundation for developing a guidance mechanism of metaheuristic algorithms for solving the VRP.
\end{abstract}

\keywords{Metaheuristic, Vehicle Routing Problem, Binary Classification, Explainable AI (XAI)}

\section{Introduction}
\label{sec:introduction}
    Combinatorial optimization problems, such as Vehicle Routing Problems (VRP), are important in real-world applications as they search for efficient solutions to minimize costs. 
    
    Despite extensive research over decades, achieving optimal solutions remains a challenge \citep{laporte2009fifty}. Furthermore, the unique constraints of various problem variants demand specialized algorithms. The development of these algorithms is complex, making Machine Learning (ML) an attractive approach to improving the existing algorithms. Routing algorithms are typically divided into two categories: exact algorithms that offer global optimum but require many computational resources and heuristics methods for practical, real-world applications that mostly find a near-optimal solution. While most heuristics rely on human-designed strategies \citep{lucas2020reducing}, ML offers a new approach improving algorithm. Moreover, the selection of features influenced by these ML models plays a critical role in effectively enhancing heuristic performances \citep{arnold2019makes,arnold2019knowledge,lucas2019comment}. Understanding the predictions of an ML model can be as crucial as the accuracy of the prediction itself in many applications \citep{lundberg2017unified}. 
    
    The necessity to interpret predictions from tree models is more common \citep{lundberg2020local}, and explaining a series of models is fundamental for debugging the algorithm design \citep{chen2022explaining}. Recent research on interpreting ML models has focused on feature attribution strategies. These methods assign a value to each input feature, indicating its importance in making a specific prediction. Feature attributions based on the Shapley value have gained favor for interpreting ML models \citep{chen2023algorithms,liu2023shapley,zern2023interventional}. The integration of ML and optimization algorithms can be classified into three mechanisms: (1) end-to-end learning, (2) learning based on problem properties, and (3) learning repeated decisions \citep{bengio2021machine}. Several studies demonstrate that learning based on problem properties and repeated decision-making is more favorable for solving larger and more realistic problems \citep{arnold2019knowledge, li2021learning, xin2021neurolkh, morabit2021machine, ma2023learning, morabit2023learning}. However, the features used in designing the learning method affect the performance of the developed hybrid algorithm.

    This work builds upon previous contributions that focused on understanding the characteristics and features of high-quality VRP solutions \citep{arnold2019makes, lucas2019comment}. In this research, we conduct a sensitivity analysis of feature importance that influences a prediction model used to evaluate VRP solution quality. The primary objective is to analyze the behavior of problem features concerning solution quality across different scenarios. Prior work has shown that understanding the structural characteristics of solutions in combinatorial optimization helps design efficient algorithms, particularly for solving the VRP \citep{arnold2019makes, arnold2019knowledge, ARNOLD201932, de2020minereduce, de2022interpretable, de2024improved, mesa2022machine, cavalcanti2023learning, santana2023neural}. While earlier studies laid the groundwork for understanding the importance of features in analyzing VRP solution quality \citep{arnold2019makes, lucas2019comment}, this study extends prior findings by incorporating multi-scenario and sensitivity analyses to uncover consistent and variable predictors of solution quality across diverse conditions.
    
    The introduction of a mechanism for ranking feature importance further differentiates this work, providing additional insight for guiding the design of metaheuristic algorithms. Our goal is to identify a set of features with high potential for formulating guidelines to enhance metaheuristic algorithms, particularly for VRP. By uncovering these important characteristic solutions, we aim to provide a foundation for developing guidance able of improving the performance and effectiveness of metaheuristic algorithms in solving VRP.

    \subsection{Research Questions and Contributions}  
    \label{subsec:research-question-and-contributions}
        Due to the $\mathcal{NP}$-Hard nature of VRP, metaheuristic methods are often used to derive solutions as they provide high-quality results within a feasible time, especially for large and complex instances \citep{Prodhon2016}. Defining optimal solution properties has been found to benefit the design of efficient algorithms \citep{arnold2019knowledge}. Explainable learning models can transform VRP solving methods \citep{lucas2019comment}, potentially substituting human-crafted strategies. Machine learning model develop predictors that map data features to classes \citep{guidotti2018survey}. By explaining these predictors, we understand feature utilization and integrate characteristic importance into rules for better guidance. In summary, this research addresses the following key questions:

        \begin{enumerate}
            \item What is the impact of each feature on the prediction accuracy of the VRP solution quality?
            \item How does the feature importance vary across different scenarios?
            \item Are there any features consistently strong predictors of solution quality regardless of the scenario?
        \end{enumerate}

        In response to these questions, a comprehensive dataset of multi-scenario VRP solutions was generated to facilitate the development of a robust learning model applicable to different scenarios. Subsequently, a multi-scenario learning model was designed to examine the behavior of the features of VRP within these scenarios. A sensitivity analysis was then performed to explore each feature's impact on the prediction model's performance. Through this analysis, we identified features that consistently influenced the predictions of the developed learning model across various scenarios. Additionally, to illustrate our findings, we proposed a novel formula to determine the order of feature importance, considering prediction accuracy across different scenarios. In summary, the main contributions of this research are:

        \begin{enumerate}
            \item A new dataset of multi-scenario VRP solutions for developing a robust learning model.
            \item A robust multi-scenario learning model that assesses the behavior of VRP features.
            \item A sensitivity analysis to determine each feature's impact on the prediction model's performance, as well as the variation and consistency of feature  importance across the scenarios.
            \item Introducing a novel formula to identify the order feature importance that considers prediction accuracy across different scenarios.
        \end{enumerate}

        In the sequel, the paper is structured as follows: \Cref{sec:methodology} details our approach for analyzing VRP features, \Cref{sec:data-mining-analysis} discusses feature behaviors and analyzes the findings, and \Cref{sec:modelling-unified} presents a unified explanation applicable across multiple scenarios.

\section{Research Methodology}
\label{sec:methodology}
\begin{figure}[H]
        \begin{center}
            \includegraphics[width=1\textwidth]{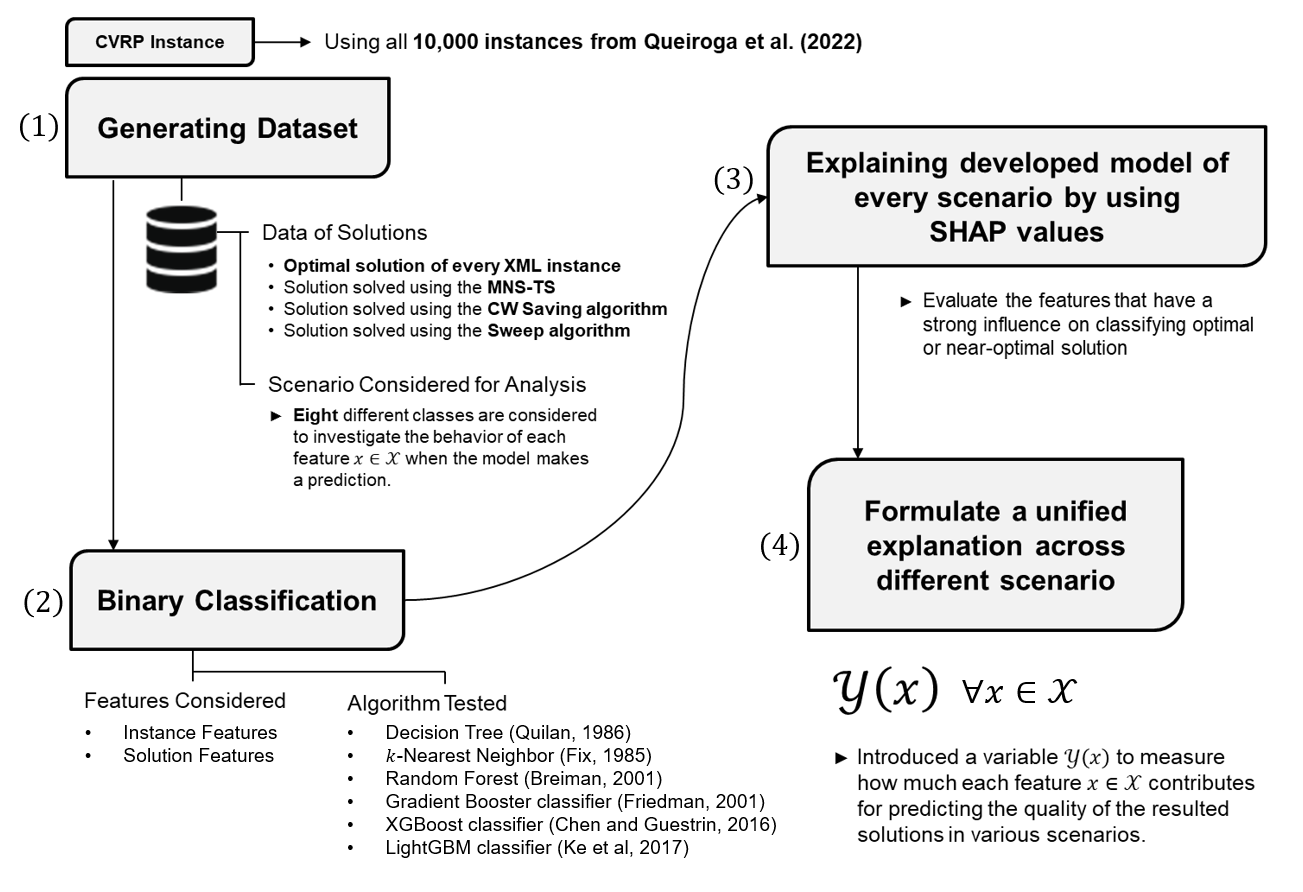}
            \caption{Research methodology used for sensitivity analysis of VRP's feature importance.}
            \label{img:pipeline}
        \end{center}
    \end{figure}

    Our methodology to conduct a sensitivity analysis of feature importance is composed of four major steps: (1) generate data, (2) build a binary classification model across multi-scenario (defined in \Cref{def:scenario}), (3) explain the model using SHAP values, and (4) formulate a unified explanation formula, as shown in \Cref{img:pipeline}. As the VRP is a complex problem with a wide variety of instances, we focus on the basic variant of VRP, which is the Capacitated Vehicle Routing Problem (CVRP). The CVRP can be characterized by an undirected graph $G=(V,E)$. The set of nodes $V$ is composed of a depot $c_0$ and a collection of customer nodes $C$, where $C=\{c_1, c_2,\cdots,c_{\mathcal{V}}\}$ and $\mathcal{V} = |V|-1$ \citep{Prodhon2016}.
    
    \begin{definition}[Scenario] \label{def:scenario}
        A scenario refers to a specific configuration of data within a classification model, particularly focusing on the distribution of the positive $(y_i = 1)$ and negative $(y_i = 0)$ classes. In this context, the positive class represents optimal solutions, while the negative class consists of near-optimal solutions with a positive gap.
    \end{definition}
    
    In this research, the objective of the learning model $\hat{f}$ is to classify between optimal and near-optimal solutions by performing a binary classification. Given a dataset $\mathcal{D} = \{ (\mathbf{x}_i, y_i)\}^N_{i=1}$, where $\mathbf{x}_i \in \mathbb{R}^d$ represents the feature vector for the $i$-th instance, and $y_i \in \{0,1\}$ is the corresponding binary label, the classifier $\hat{f}$ can be represented as $\hat{f}: \mathcal{X} \rightarrow \{0, 1\}$, where $\mathcal{X}= \mathbb{R}^d$ is the instance space and the $d$ denotes the number of features. The $N$ represents the total number of solution in the dataset $\mathcal{D}$, where each $\mathbf{x}_i$ being a $d$-dimensional feature vector. The features $d$ used in developing the learning model are described in \Cref{subsec:feature-used}.
    
    \subsection{Problem Instances Used and Optimal Solutions}
    \label{subsec:optimal}
        The $\mathbb{XML}$100 instances\footnote{All the information related to the instances and their optimal solutions are available at \url{http://vrp.galgos.inf.puc-rio.br/index.php/en/}.} are employed to generate solution data, which is subsequently used for analysis to comprehend feature behaviour and develop classification model $f(\mathbf{x})$. The 10,000 $\mathbb{XML}$100 instances consist of a similar number of customer nodes, totalling 100 nodes each. However, they vary across different categories, such as the position of their depots, the distribution of customers, the distribution of demands, and the average size of the routes in their optimal solutions \citep{queiroga202110}.

    \subsection{Data of Near-Optimal Solutions}
    \label{subsec:near-optimal-solutions-all}
        To achieve binary classification between optimal solutions and near-optimal solutions using the model $\hat{f}$, we utilized the optimal solutions from every instance of the $\mathbb{XML}$100 instances. To generate near-optimal solutions, we employed the MNS-TS algorithm \citep{soto2017multiple,lucas2020reducing}, the Clarke-Wright Savings algorithm \citep{clarke1964scheduling}, and the Sweep algorithm \citep{gillett1974heuristic} on all $\mathbb{XML}$100 instances\footnote{The Clarke-Wright and Sweep algorithm were executed by using solver by \citet{groer2010library}, available at \url{https://github.com/coin-or/VRPH}.}. The MNS-TS algorithm, which stands for \textit{Multiple Neighborhood Search with Tabu Search}, has proven effective in solving both the Open VRP (OVRP) \citep{soto2017multiple} and large-scale CVRP \citep{lucas2020reducing}. 

        \begin{definition}[Gap to Optimal] \label{def:gap-to-bks}
            Given $\text{Obj}(\mathcal{Z}_1)$ as the total distance of optimal solution and $\text{Obj}(\mathcal{Z}_0)$ as the total distance of near-optimal solution, the relative difference between the objective value of a near-optimal solution, as well as the value of the optimal solution are expressed as the gap to optimal, calculated as:
            \begin{equation} \label{eq:gap-to-bks}
                \text{Gap to Optimal} := \dfrac{\text{Obj}(\mathcal{Z}_0) - \text{Obj}(\mathcal{Z}_1)}{\text{Obj}(\mathcal{Z}_1)} \times 100\%
            \end{equation}
        \end{definition}

        \begin{figure}[H]
            \begin{center}
                \includegraphics[width=0.6\textwidth]{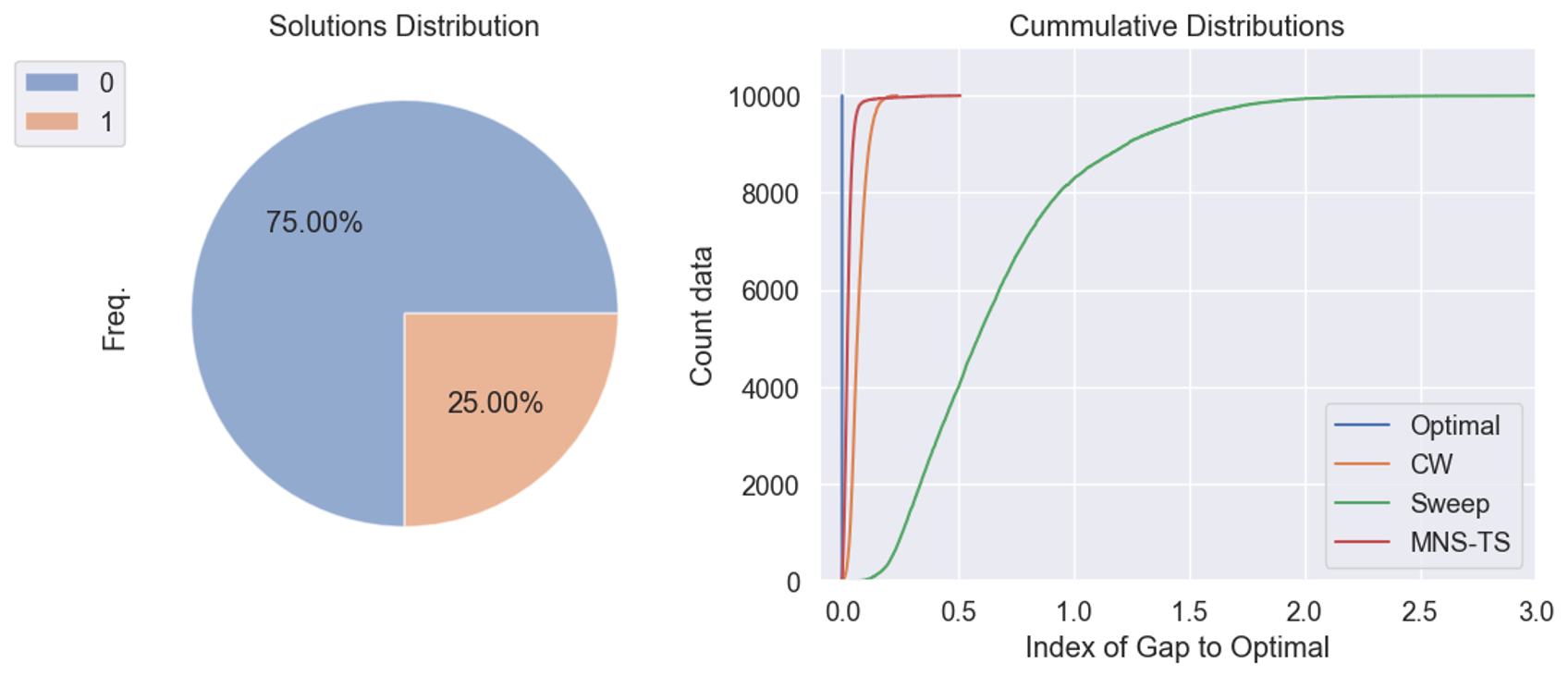}
                \caption{The pie chart (left) shows the dataset split into optimal $(y_i=1)$ and near-optimal $(y_i=0)$ solutions. The cumulative chart (right) shows the distribution of near-optimal solutions, where Sweep offers a wider range of solution qualities that the others.}
                \label{img:sol-distribution}
            \end{center}
        \end{figure}        
        Given \Cref{def:scenario}, for the analysis, the near-optimal solutions are divided into eight groups, each corresponding to a different scenario. These scenarios, along with their respective near-optimal solutions, are detailed in \Cref{table:scenario-setting}. These scenarios allow for the analysis of feature importance and model behavior under varying conditions of data distribution, particularly with respect to \Cref{def:gap-to-bks}. 

    \begin{table}[htbp]
        \centering
        \scalebox{0.86}
        {
        \begin{threeparttable}
            \caption{Detailed near-optimal solution for every scenario.}
            \label{table:scenario-setting}
            \vspace*{0.1cm}
            \begin{tabular}{lcl}
                \toprule
                \textbf{Scenario}&& \textbf{Near-Optimal Solution Used for Analysis}\\
                \midrule
                \midrule
                Scenario 1: $S_1$&& Solutions solved by the MNS-TS\\
                \midrule
                Scenario 2: $S_2$&& Solutions solved by the CW Saving algorithm\\
                \midrule
                Scenario 3: $S_3$&& Solutions solved by the Sweep algorithm\\
                \midrule
                Scenario 4: $S_4$&& All solutions with gap $> 2\%$\\
                \midrule
                Scenario 5: $S_5$&& All solutions with gap $> 5\%$\\
                \midrule
                Scenario 6: $S_6$&& All solutions with gap $> 7\%$\\
                \midrule
                Scenario 7: $S_7$&& All solutions with gap $> 10\%$\\
                \midrule
                Scenario 8: $S_8$&& All solutions with gap $> 15\%$\\
                \bottomrule
            \end{tabular}
            \begin{tablenotes}
                \item Note: all scenarios include optimal solutions for comparison.
            \end{tablenotes}
        \end{threeparttable}
        }
    \end{table}

    \subsection{Features of Problem}
    \label{subsec:feature-used}
        This research categorizes the features for developing the learning model into instance-based and solution-based features. Some of these features are derived from the mean value or standard deviation (SD) of certain characteristics. The features used in this research are based on findings from prior recent studies \citep{herdianto:hal-04310676}. The instance-based and solution-based features are summarized in \Cref{table:feature-used}. In total, there are 9 features related to the instance characteristics and 22 features related to the characteristics of the resulting solution.
        
        \begin{table}[htbp] 
            \caption{Complete list of features used for analysis.}
            \label{table:feature-used}
            \vspace*{0.1cm}
            \centering
            \scalebox{0.8}
            {
                \begin{tabular}{lll}
                \toprule
                    \textbf{Instance-based Features} & \multicolumn{2}{l}{\textbf{Solution-based Features}}\\
                \midrule
                \midrule
                    \textbf{I01:} Number of customers & \textbf{S01:} Mean width of each route& \textbf{S12:} Mean terminal demand customer\\
                    \textbf{I02:} Number of vehicles & \textbf{S02:} SD the width of each route&  \textbf{S13:} Mean the farthest customer's demand\\
                    \textbf{I03:} Degree of capacity utilization & \textbf{S03:} Mean span of each route& \textbf{S14:} SD the farthest customer's demand\\
                    \textbf{I04:} Mean customer-paired edge & \textbf{S04:} SD the span of each route& \textbf{S15:} SD the length of each route\\
                    \textbf{I05:} SD customer-paired edge & \textbf{S05:} Mean depth of each route& \textbf{S16:} Mean route-centroid distances\\
                    \textbf{I06:} Mean customer-depot edge & \textbf{S06:} SD the depth of each route&  \textbf{S17:} SD route-centroid distances\\
                    \textbf{I07:} SD the customer-depot edge & \textbf{S07:} Route-edge ratio& \textbf{S18:} Mean degree of neighborhood\\
                    \textbf{I08:} Mean angular deviation of customers & \textbf{S08:} Mean max edge& \textbf{S19:} Mean route's capacity utilization\\
                    \textbf{I09:} SD angular deviation of customers & \textbf{S09:} Longest-edge ratio& \textbf{S20:} SD route's capacity utilization\\
                    & \textbf{S10:} longest-edge ratio &  \textbf{S21:} Mean of longest distance-relatedness\\
                    & \textbf{S11:} Mean terminal edge& \textbf{S22:} SD of longest distance-relatedness\\
                \bottomrule
                \end{tabular}
            }
        \end{table}

    \subsection{Classifier Baselines}
    \label{subsec:clf}
        For analyzing the behavior of features when predicting solutions, we developed the classifier model $\hat{f}$ using several baseline classifiers, including $k$-Nearest Neighbors ($k$NN) \citep{fix1985discriminatory}, Decision Tree classifier \citep{quinlan1986induction}, and Random Forest \citep{breiman2001random}. $k$NN is a simple, non-parametric classifier that classifies a data point based on the majority class of its $k$-nearest neighbors. Meanwhile, the Decision Tree classifier recursively splits the feature space to form a tree structure, making it easy to interpret but prone to overfitting. Random Forest addressed the mechanism of decision trees by combining multiple trees trained on random subsets of data and features. We also utilized various boosting algorithms, such as Gradient Boosting \citep{friedman2001greedy}, Extreme Gradient Boosting (XGBoost) \citep{chen2016xgboost}, and Light Gradient Boosting Machine (LightGBM) \citep{ke2017lightgbm}. Gradient Boosting is an ensemble method that builds decision trees sequentially, with each tree correcting the errors of the previous ones by optimizing a loss function. XGBoost refines the boosting mechanism by incorporating grow trees level-wise approach to reduce overfitting. LightGBM further refines this by using a histogram-based approach to find splits and grow trees leaf-wise (as opposed to XGBoost's level-wise approach).
        
    \subsection{Metric Evaluation}
    \label{subsec:metric-eval-f1}
        The evaluation of the classifier $\hat{f}$ is based on the confusion matrix. As outlined by \citep{narasimhan2014statistical}, the confusion matrix provides a tabular summary of the classification performance for a given dataset $\mathcal{D}$, comprises True Positives (TP), False Positives (FP), True Negatives (TN), and False Negatives (FN). The confusion matrix $\mathcal{C}$ can be represent as a $2$-dimensional matrix, as:
        \begin{equation} \label{eq:confusion-matrix}
        \mathcal{C} := \begin{bmatrix}
                            \text{TN} & \text{FP} \\
                            \text{FN} & \text{TP}
                       \end{bmatrix}
        \end{equation} 
        Given a threshold of 0.5 for the classifier $\hat{f}$, the classification rule is defined as follows: if $\eta(\mathbf{x}) < 0.5$, the prediction is $\hat{f}(\mathbf{x}) = 0$ (considered a near-optimal solution), and if $\eta(\mathbf{x}) \geq 0.5$, the prediction is $\hat{f}(\mathbf{x}) = 1$ (considered an optimal solution). Accordingly, the key components of $\mathcal{C}$ in \Cref{eq:confusion-matrix} can be defined: TP as $\text{TP} := \mathbb{P} \left( \hat{f}(\mathbf{x}) = 1 \land y = 1 \right)$, and TN as $\text{TN} := \mathbb{P} \left( \hat{f}(\mathbf{x}) = 0 \land y = 0 \right)$. Similarly, FP is defined as $\text{FP} := \mathbb{P} \left( \hat{f}(\mathbf{x}) = 1 \land y = 0 \right)$, and FN as $\text{FN} := \mathbb{P} \left( \hat{f}(\mathbf{x}) = 0 \land y = 1 \right)$. Various metrics can be derived from the confusion matrix $\mathcal{C}$ to evaluate the performance of $\hat{f}$, one of which is the Precision-Recall (PR) curve \citep{davis2006relationship}.

        \begin{equation} \label{eq:recall}
            \text{recall} = \mathbb{P} \left( \hat{f}(\mathbf{x}) = 1 | y = 1 \right) = \dfrac{\mathbb{P} \left( \hat{f}(\mathbf{x})=1 \land y=1\right) }{\mathbb{P}(y=1)} = \dfrac{\text{TP}}{\text{TP}+\text{FN}}
        \end{equation}
        \begin{equation} \label{eq:precision}
            \text{precision} = \mathbb{P} \left( y = 1 | \hat{f}(\mathbf{x}) = 1 \right) = \dfrac{\mathbb{P} \left( \hat{f}(\mathbf{x})=1 \land y=1\right) }{\mathbb{P}(\hat{f}(\mathbf{x})=1)} = \dfrac{\text{TP}}{\text{TP}+\text{FP}} 
        \end{equation}
        Given the confusion matrix \Cref{eq:confusion-matrix}, precision measures the accuracy of positive predictions, indicating the proportion of instances classified as positive that are actually positive. Meanwhile recall measures the proportion of actual positives correctly identified by the classifier \citep{narasimhan2014statistical}. Thus, $F_{\beta}$-score is used for quantifying the trade-off between precision and recall \citep{sokolova2009systematic}. The $F_{\beta}$-score is advantageous for evaluating binary classifiers, in particular with imbalanced data, because it balances precision and recall. In cases where one class is much more common than the other, metrics like accuracy can be misleading, as a classifier might achieve high accuracy by simply predicting the majority class. The $F_{\beta}$-score solve this issue by considering both FP and FN, making it a better performance indicator on the minority class.

        \begin{definition}[$F_{\beta}$-score] \label{def:f-score}
            The $F_{\beta}$-score is defined as weighted average of $\text{precision}$ and $\text{recall}$, calculated as:
            \begin{equation} \label{eq:f-score}
                F_{\beta}\text{-score} := \dfrac{(\beta^2 + 1) \cdot \text{precision} \cdot \text{recall}}{\beta^2 \cdot \text{precision} + \text{recall}}
            \end{equation}
            where $\beta$ is a variable that controls the importance of $\text{recall}$ in the overall score.
        \end{definition}
        
        Given \Cref{def:f-score}, by substituting $\beta = 1$ into \Cref{eq:f-score}, the $F_{\beta}$-score provides a balanced measure of $\text{precision}$ and $\text{recall}$, known as the $F_1$-score. Thus, in this research will utilize the $F_1$-score as the key metric to evaluate the classifier's performance in every scenario as it able for ensuring the classifier $\hat{f}$ perform well with respect to the positive class $y_i=1$ (optimal solutions).
        
    \subsection{Model Interpretability} 
    \label{subsec:interpretability}
        Model interpretability is the ability to interpret and understand how a machine learning model able to generates its predictions and decisions \citep{guidotti2018survey}. Model interpretability can be achieved by providing understandable explanations for every predictions. Various methods exist to achieve model explainability, one of them is the SHAP value (Lundberg et al., \citeyear{lundberg2017unified, lundberg2020local}; Baptista et al., \citeyear{baptista2022relation}). SHAP values quantify the contribution of each feature by assessing the effect of its inclusion in every possible subset of the remaining features, thereby capturing feature interactions.
        \begin{definition}[SHAP (SHapley Additive exPlanations)] \label{def:shapely}
            Let $\phi_j$ represents the SHAP value for feature $j$, reflecting its average contribution to the model's output across all possible feature combinations. Assume $\mathbf{D}$ as the set of all features, where the $d$ represents $|\mathbf{D}|$. Then, the $\mathbf{U}$ is a subset of $\mathbf{D}$ that does not include $j$, where the $u$ represents the size of $\mathbf{U}$. Thus, assume $v(\mathbf{U})$ is the value function, which typically represents the model's prediction when only features in set $\mathbf{U}$ are considered. For interpreting ML models, the SHAP value for feature $j$ is computed by considering all possible subsets of features excluding $j$ and the change in the prediction of the classifier when $j$ is added to the subset. The SHAP value for feature $j$ is given by:
            \begin{equation} \label{eq:shapley}
                \phi_j := \sum_{\mathbf{U} \subseteq \mathbf{D} \setminus \{j\}} \frac{u!(d-u-1)!}{d!} \left[ v(\mathbf{U}\cup \{j\}) - v(\mathbf{U}) \right]
            \end{equation}
        \end{definition}
        
        Hence, given by \Cref{def:shapely}, we can know that a higher absolute mean SHAP value of a feature indicates a stronger influence of that feature on the model's predictions. Moreover, SHAP values also provide several advantages, including interaction awareness. SHAP values capture feature interactions by calculating each feature's contribution to predictions, offering a comprehensive understanding of feature importance \citep{lundberg2017unified}.
        
\section{Data Mining Analysis}
\label{sec:data-mining-analysis}
    As summarized in \Cref{img:pipeline}, several classifiers were performed on the dataset for each scenario. The best classifier was then further analyzed to obtain its SHAP values\footnote{The dataset, the source code and the documentation for performing the analysis can be found at \url{https://github.com/bachtiarherdianto/Modelling-XAI-CVRP}.}, providing insights into the contribution and impact of each feature on the model's predictions.

    \begin{figure}[H]
        \begin{center}
            \includegraphics[width=0.99\textwidth]{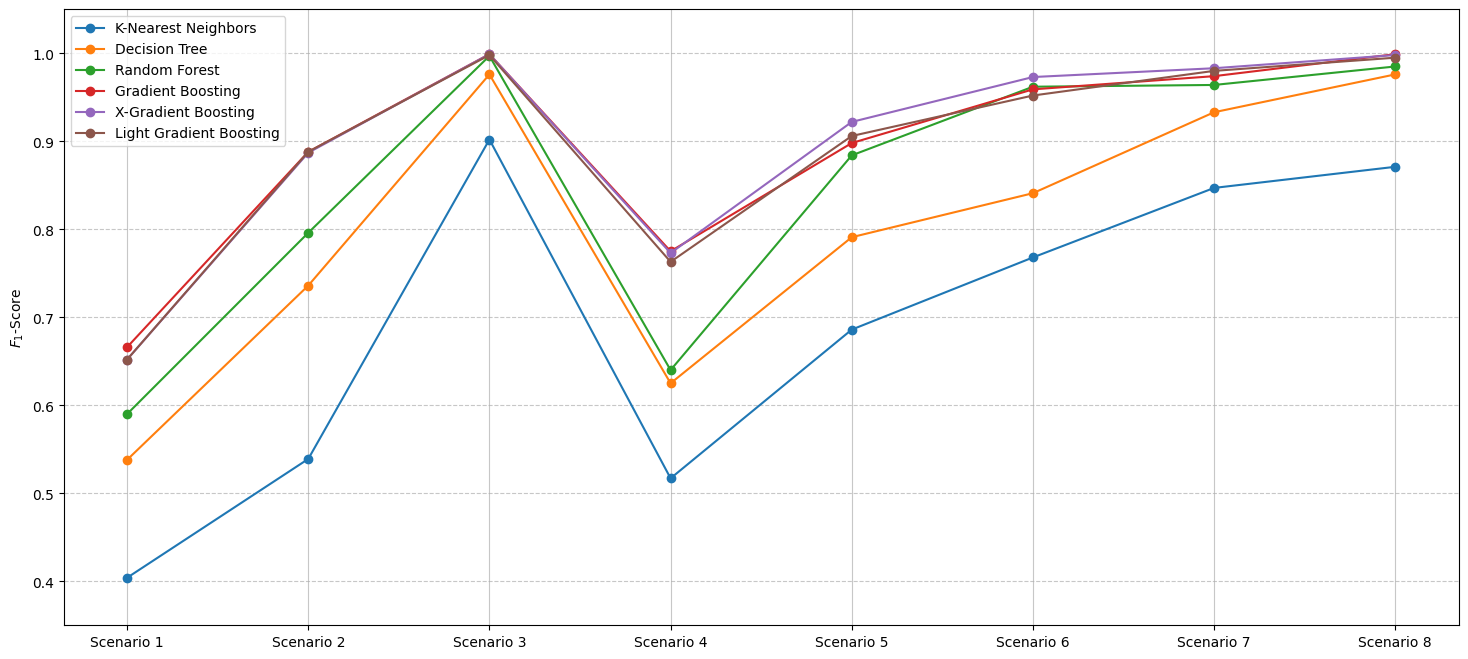}
            \caption{Performance of classifiers in every scenario.}
            \label{img:clf-summary}
        \end{center}
    \end{figure}

\begin{figure}[H]
        \begin{center} 
            \includegraphics[width=1\textwidth]{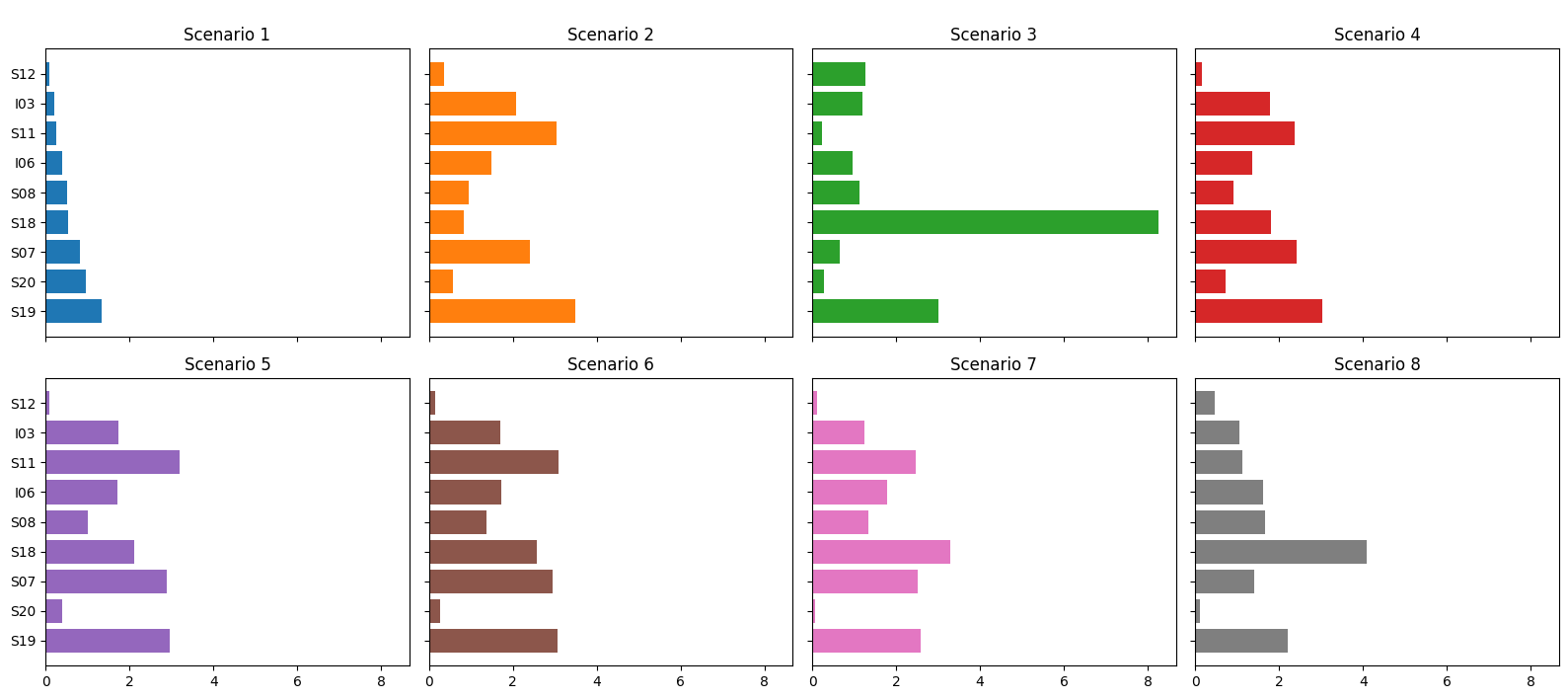}
            \caption{SHAP value of filtered features in every scenario.}
            \label{img:shap-summary}
        \end{center}
    \end{figure}

    In \Cref{img:clf-summary}, the performance of various classifiers across 8 scenarios is summarized. Meanwhile \Cref{img:shap-summary} shows the global feature importance based on SHAP values of filtered features. In the first 3 scenarios, individual heuristic algorithms were used to generate solutions. Gradient Boosting and X-Gradient Boosting classifiers consistently achieved the highest $F_1$-scores, while $k$-NN lagged behind. Features like S19 and S18, representing the average rank of the neighborhood and capacity utilization of the VRP, respectively, were consistently identified as the most important across all scenarios, underscoring their important role in model prediction. In scenarios four through eight, solutions from all 3 algorithms were combined, focusing on those with increasing gaps from the optimal solutions to be more diversified solutions. The performance of the classifiers generally improved as the quality solution diversified, a similar trend to the previous first 3 scenarios. This trend is further explained in \Cref{prop:gap-to-f1}. Meanwhile, in \Cref{img:shap-summary}, each scenario demonstrated unique distributions of feature significance, with some scenarios showing one or two dominant features while others exhibited a more balanced spread. For instance, Scenario 3 had a clear standout feature (S07), whereas in Scenario 5, the distribution of importance was more even among several features like I03, S11, and S18. These variations suggested the dynamic nature of feature importance across different models, datasets, or conditions, where feature significance could shift depending on the specific scenario.

    \begin{proposition}\label{prop:gap-to-f1} 
        Let $\mathcal{Z}_1^m$ represent the data points $y_i = 1$ (optimal solutions) in scenario $m \in M$, and $\mathcal{Z}_0^m$ as $y_i = 0$ (near-optimal solutions). If a dataset of a scenario $m$ exhibits a wide variety in the distribution of data points such that the $\text{gap}$ between $\mathcal{Z}_1^m$ and $\mathcal{Z}_0^m$ are diverse, then the classifier $\hat{f}$ will perform better, as measured by the $F_1$-score.
    \end{proposition}
    \begin{proof}\label{proof:gap-to-f1}
        Given the formulation of $\text{gap to optimal}$ in \Cref{eq:gap-to-bks} and $F_1$-score in \Cref{eq:f-score} when $\beta=1$, we can conclude that with better separation and fewer miss-classifications between $\mathcal{Z}_1$ and $\mathcal{Z}_0$, precision improves as $\hat{f}$ has less FP. Moreover, a diverse dataset helps the classifier $\hat{f}$ improves recall by reducing FN.
    \end{proof}

\begin{figure}[H]
        \begin{center} 
            \includegraphics[width=1\textwidth]{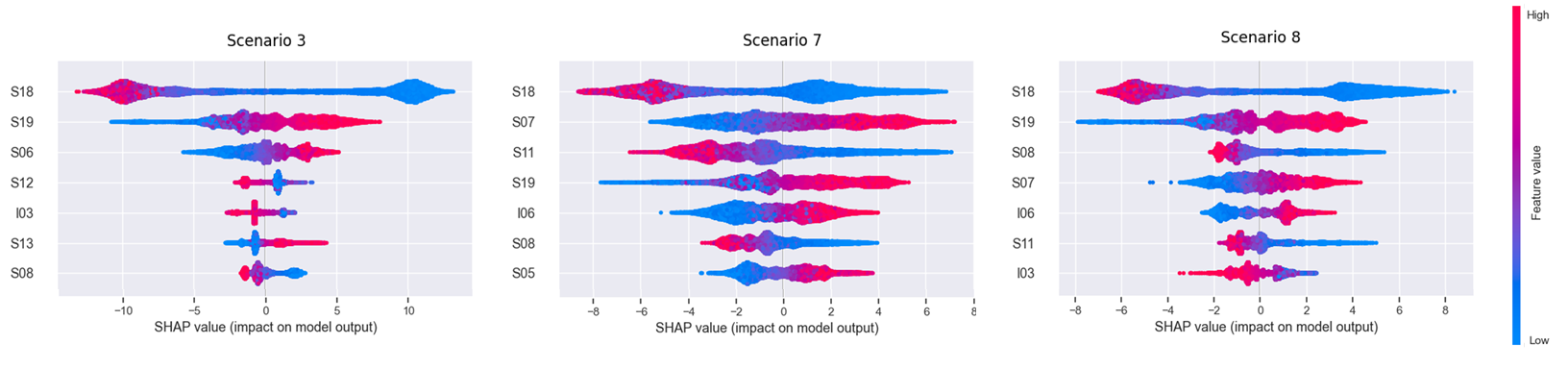}
            \caption{The local explanation summary plot for filtered features from selected scenarios.}
            \label{img:shap-beeswarm-summary}
        \end{center}
    \end{figure}

    \Cref{img:shap-beeswarm-summary} illustrate the SHAP values of several strong features in three selected highest $F_1$-score scenarios. Across all selected scenarios, S18 and S19 repeatedly emerge as the most importance features. The color gradient in the plot represents the behavior of the features, where higher values of S19 (indicated by red points) are often associated with positive SHAP values, which correlate with better solution quality. Conversely, lower values of S18 (blue points) could also correspond to better solution quality, as indicated by their positive SHAP values. This suggests that while S18 and S19 are consistently important, their impact on solution quality can vary depending on whether their values are high or low. Moreover, given by \Cref{def:scenario} and \Cref{eq:shapley}, we can conclude that the directional impact of SHAP values for any given feature will mostly similar across different scenarios. However, the order of feature importance can vary between scenarios. This is because the direction of a feature's SHAP value $\phi_j$ is mostly preserved because the scalar relationship between the feature and the target variable mostly similar. Meanwhile, the absolute values can differ as the influences of features across scenarios vary, leading to the variation order of feature importance.
    
    \ref{sec:appendix-detailed-explanation} shows the details of the global feature importance plots and the local explanation summary plot for every scenario. The global feature importance plot can be used to identify which features are most important in the decision-making process, while the local explanation summary plot is useful for understanding the behaviour of features when the model makes a prediction. 
    
\section{Modelling Across Multi-Scenario}
\label{sec:modelling-unified}
    As shown in \Cref{img:shap-summary} and \ref{img:shap-beeswarm-summary}, features S18 and S19 frequently emerge as the most important features in several scenarios. This is happens particularly in scenario 3, indicating their significant contribution in characterizing solution quality. However, this dominance is only sometimes apparent, as shown in \Cref{img:shap-summary}. The other features, such as S11, S07, and I03, also exhibit significant influence in several other scenarios, making it seem that they might be equally influential. The apparent competitiveness of these other features can create the impression that S18 and S19 are not overwhelmingly strong influences $\hat{f}$. Nevertheless, when considering the $F_1$-score associated with each scenario, the actual effect of S18 and S19 becomes clearer. With factoring in the $F_1$-scores, the comparative strength of S18 and S19 might seem more convincing. Given this analysis, \Cref{def:unified-shap} explain how to visualize in a simple manner the feature importance rank across different tested scenarios.
    
    \begin{definition}\label{def:unified-shap} 
        Given the SHAP value $S^m(x)$ of a feature $x$ in scenario $m$ and the $F_1\text{-score}_m$ of the classifier in that scenario, the overall contribution $\mathcal{Y}(x)$ of feature $x$ to the quality of solutions across multiple scenarios is defined as:
        \begin{equation} \label{eq:unified-y}
            \mathcal{Y}(x) = \sum_{m \in M} \left( S^m(x) \times F_1\text{-score}_m \right)
        \end{equation}
        where $M$ is the set of all scenarios, and $S^m(x)$ is the mean of the absolute SHAP values of feature $x$ in scenario $m$.
    \end{definition}
    
    From \Cref{eq:unified-y}, we can see that the SHAP value $S^m(x)$ quantifies the impact of a feature on model predictions in a given scenario $m \in M$, where $M = 8$ (as summarized in \Cref{table:scenario-setting}). Meanwhile, The $F_1$-score shows the model's accuracy. To determine the overall contribution $\mathcal{Y}(x)$ of a feature across all scenarios, we weight its SHAP values by the respective scenario's $F_1$-score and sum them across all scenarios. Given \Cref{def:f-score} and \ref{def:shapely}, \Cref{eq:unified-y} captures the cumulative influence of a feature by considering its impact on model's predictions and the model’s performance.

    \begin{figure}[H]
        \begin{center}
            \includegraphics[width=1\textwidth]{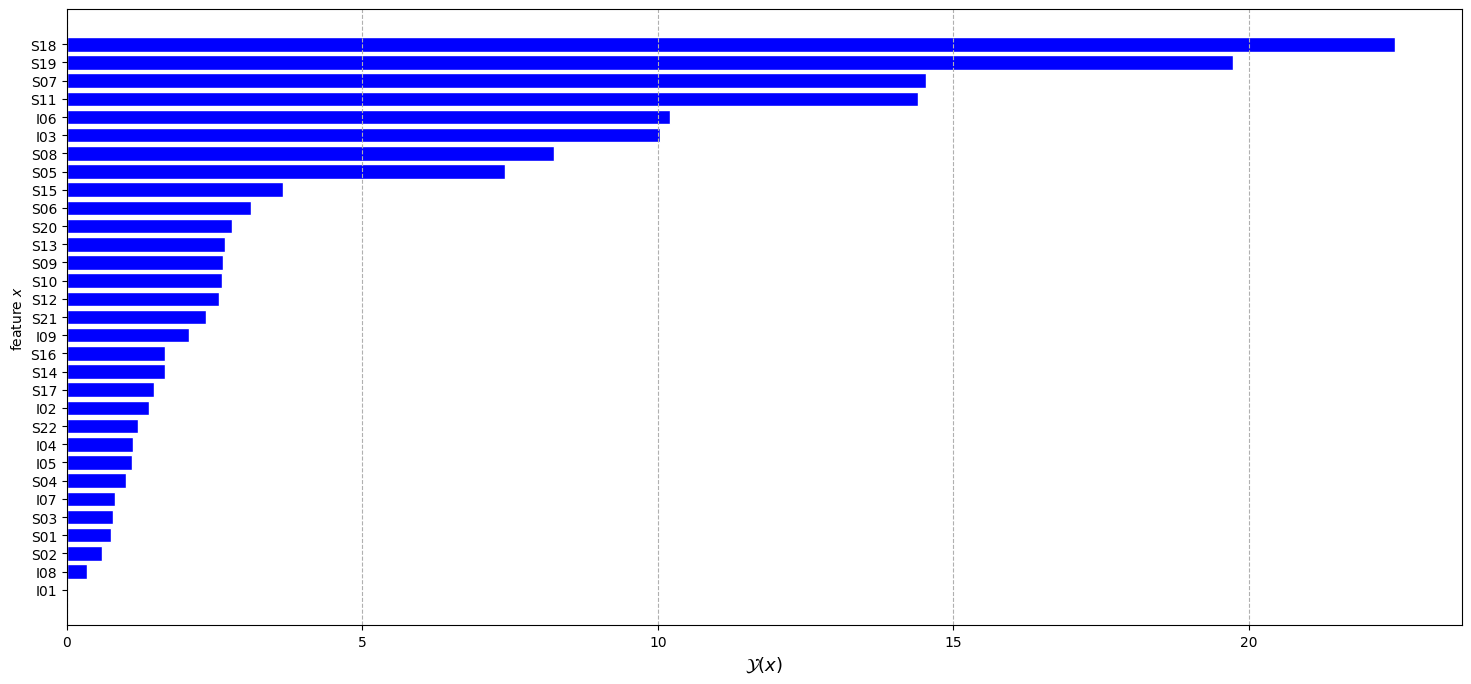}
            \caption{The representation of the VRP's feature importance across multiple tested scenarios.}
            \label{img:beeswarm-combi}
        \end{center}
    \end{figure}

    \Cref{img:beeswarm-combi} illustrates the relative importance of various features across multiple scenarios, recognizing S18 and S19 as the most significant. These features consistently contribute significantly to the model's prediction. S07, S11, S11, I06, and I03 (summarized in \Cref{table:feature-used}) are also important but have a lower impact compared to the previous features. Meanwhile, features like S09, S10, S12, and S21 show lower importance, with I08 and I01 being the weakest across the scenarios. Then, from \Cref{img:beeswarm-combi}, we can see that if a feature $x$ has a very high SHAP value $S^m(x)$ in a single scenario $m \in M$, and the scenario also has a relatively high $F_1\text{-score}_m$, compared to other scenarios, then feature $x$ will still maintain significant importance across different scenarios as reflected in its overall contribution $\mathcal{Y}(x)$. Meanwhile, from \Cref{img:beeswarm-combi}, we also see that if a feature $x$ consistently has high SHAP values $S^m(x)$ across multiple scenarios where the $F_1$-score is also high, then $\mathcal{Y}(x)$ will be significantly higher than other features. This implies that feature $x$ is critically important to the model’s predictive performance across those scenarios.
    
    This comprehensive understanding of feature importance provides a clearer direction of how the feature behaves when deciding. The analysis highlights the significance of S18 and S19 as key predictors, underlines the consistent influence of S19 and S18 on model predictions across different scenarios. Unlike prior studies that emphasized average route width as the main characteristic \citep{arnold2019makes, lucas2019comment}, our analysis across multiple scenarios identified a new and more robust characteristic for indicating solution quality. Subsequently, since $F_1$-score is a positive scalar, given $\mathcal{Y}(x)$ from \Cref{eq:unified-y}, the SHAP's directional behavior of a feature's influence on classifier model predictions is preserved. This because from \Cref{eq:shapley}, when multiplying the SHAP value $\phi(x)$ of a feature $x$ in a scenario $m$ by a scalar value $F_1\text{-score}_m$, the direction of $\phi(x)$ does not change.  Based on this fact, we can more precisely define the relationship between these features and solution quality. In particular, higher values of S19 consistently emerge as an indicators of better solution quality with lower $\text{gap}$ (as descripted in \Cref{def:gap-to-bks}), suggesting that as S19 increases, the likelihood of achieving a lower $\text{gap}$ of a solution. Conversely, lower values of S18 appear to correspond with improved solution quality, indicating that when S18 decreases, the model is more likely to generate a lower $\text{gap}$ of a solution. 

\newpage
\section{Conclusion}
\label{sec:conclusion}
    In this research, we investigated the influence and variability of individual features on predicting the solution quality of the Vehicle Routing Problem (VRP). We examined how each feature impacts model performance and how the importance of these features varies across different scenarios. By constructing a comprehensive dataset encompassing multiple scenarios, we developed a robust multi-scenario learning model capable of interpreting VRP feature behavior under varied conditions. Sensitivity analysis was employed to quantify each feature's impact, revealing several features that consistently influenced model performance. Additionally, we observed that a more diverse gap among near-optimal solutions enhances classification accuracy, as better separation leads to fewer false positives and reduced false negatives. To support our findings, we introduced a new formula for determining feature importance, incorporating prediction accuracy across scenarios. This work advances the understanding of feature behavior in predictive models and enhances the interpretability of VRP solution quality predictions beyond prior studies.

    While this study provides valuable insights, it represents only a starting point toward developing feature-based guidance mechanisms. The derived rules offer the potential for more precise adjustments of methods and settings based on the specific needs of each situation within the heuristic process. This opens the possibility of developing a robust guidance mechanism within metaheuristic algorithms for solving the VRP.

\newpage
\appendix
\section{Appendix: The Global Feature Importance (left) and Local Explanation Summary (right) Plots  for Every Scenario}
\label{sec:appendix-detailed-explanation}

\subsection{Scenario 1}
\label{subsec:app-scr1}

\begin{table}[htbp]
\caption{Performance of several baseline classifier used in the scenario 1.}
\label{table:f1-scr1}
\vspace*{0.1cm}
\centering
\scalebox{0.88}
{
    \begin{tabular}{lcccc}
        \toprule
            \textbf{Algorithm}                      & \textbf{Precision}  & \textbf{Recall}   && \textbf{$F_1$-score}\\
        \midrule
            K-Nearnest Neighbors Classifier         & 0.476               & 0.351             && 0.404\\
            Decision Tree Classifier                & 0.537               & 0.538             && 0.538\\
            Random Forest Classifier                & 0.523               & 0.500             && 0.511\\
            \textbf{Gradient Boosting Classifier}    & \textbf{0.672}      & \textbf{0.661}    && \textbf{0.666}\\
            X-Gradient Boosting Classifier          & 0.632               & 0.621             && 0.627\\
            Light Gradient Boosting Classifier      & 0.652               & 0.652             && 0.652\\
        \bottomrule
    \end{tabular}
}
\end{table}

\begin{figure}[H]
    \begin{center}
        \includegraphics[width=1\textwidth]{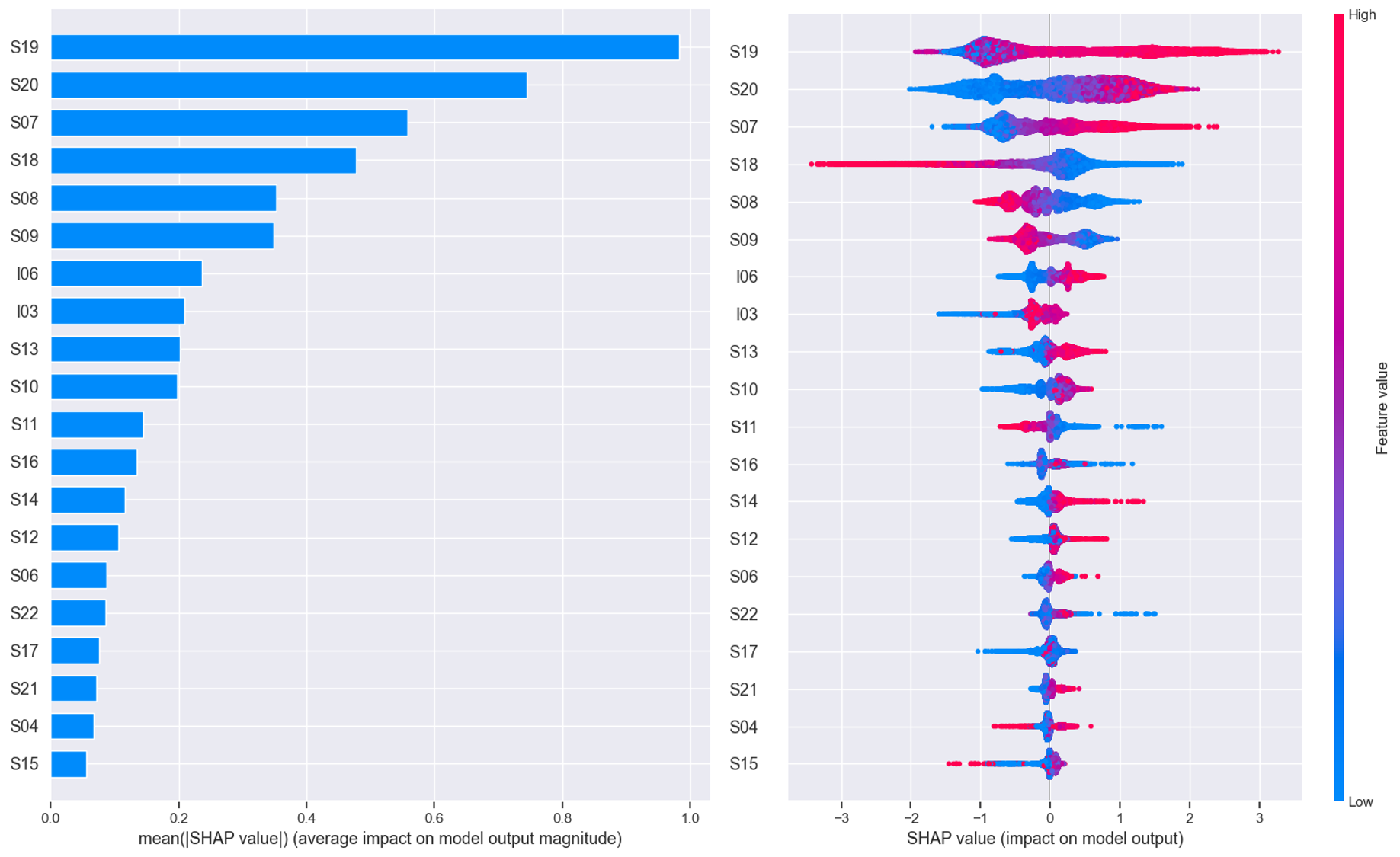}
        \caption{Plot of scenario 1 by using Gradient Boosting classifier with $F_1$-score $= 0.666$.}
        \label{img:app-1}
    \end{center}
\end{figure}

\newpage
\subsection{Scenario 2}
\label{subsec:app-scr2}

\begin{table}[htbp]
\caption{Performance of several baseline classifier used in the scenario 2.}
\label{table:f1-scr2}
\vspace*{0.1cm}
\centering
\scalebox{0.88}
{
    \begin{tabular}{lcccc}
        \toprule
            \textbf{Algorithm}                      & \textbf{Precision}  & \textbf{Recall}   && \textbf{$F_1$-score}\\
        \midrule
            K-Nearest Neighbors Classifier         & 0.609               & 0.483             && 0.539\\
            Decision Tree Classifier                & 0.742               & 0.731             && 0.736\\
            Random Forest Classifier                & 0.785               & 0.807             && 0.796\\
            Gradient Boosting Classifier            & 0.860               & 0.905             && 0.882\\
            \textbf{X-Gradient Boosting Classifier} & \textbf{0.870}      & \textbf{0.907}    && \textbf{0.888}\\
            Light Gradient Boosting Classifier      & 0.854               & 0.900             && 0.876\\
        \bottomrule
    \end{tabular}
}
\end{table}

\begin{figure}[H]
    \begin{center}
        \includegraphics[width=1\textwidth]{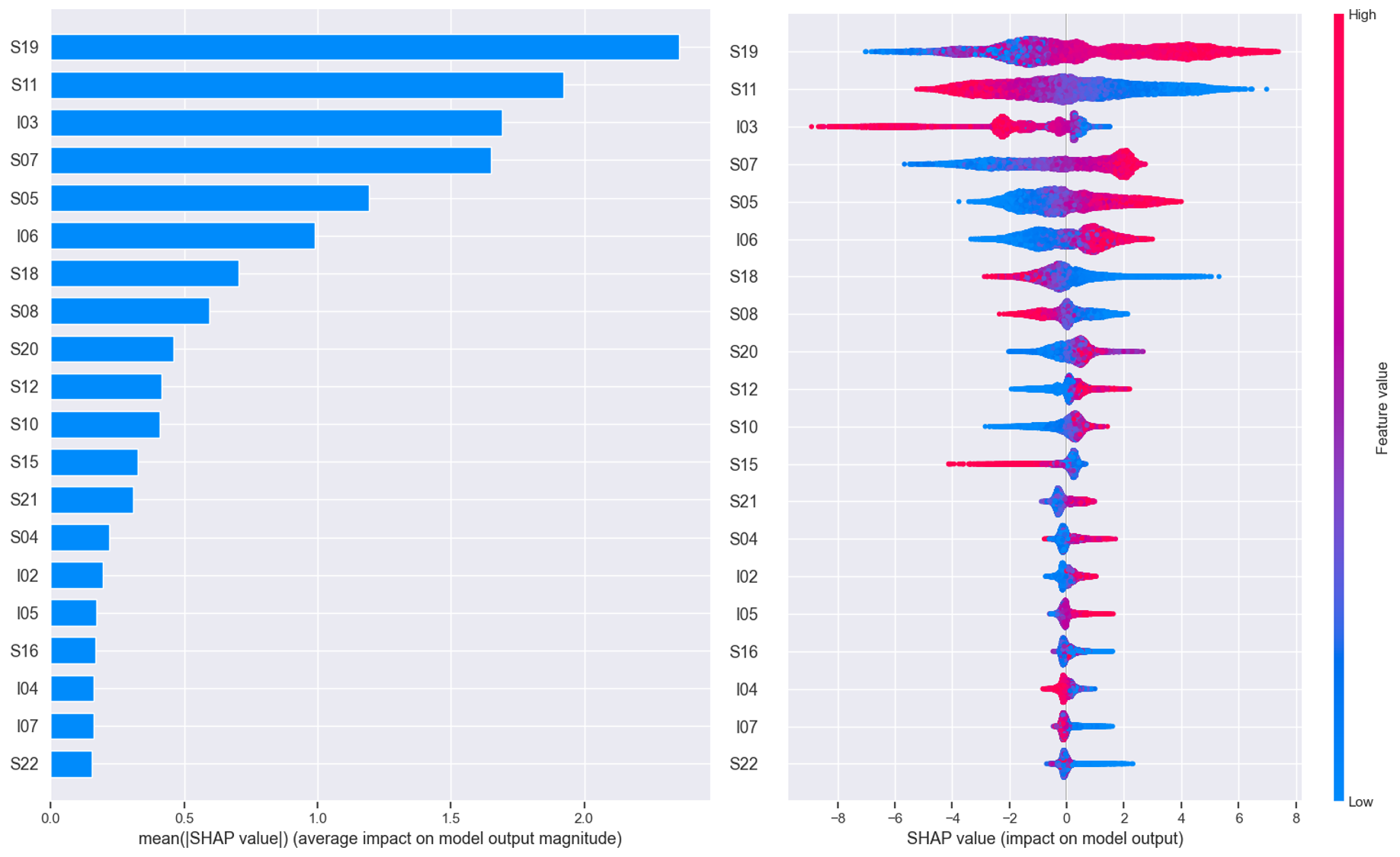}
        \caption{Plot of scenario 2 by using X-Gradient Boosting classifier with $F_1$-score $= 0.888$.}
        \label{img:app-2}
    \end{center}
\end{figure}

\newpage
\subsection{Scenario 3}
\label{subsec:app-scr3}

\begin{table}[htbp]
\caption{Performance of several baseline classifier used in the scenario 3.}
\label{table:f1-scr3}
\vspace*{0.1cm}
\centering
\scalebox{0.88}
{
    \begin{tabular}{lcccc}
        \toprule
            \textbf{Algorithm}                      & \textbf{Precision}  & \textbf{Recall}   && \textbf{$F_1$-score}\\
        \midrule
            K-Nearnest Neighbors Classifier         & 0.913               & 0.891             && 0.902\\
            Decision Tree Classifier                & 0.990               & 0.992             && 0.990\\
            Random Forest Classifier                & 0.999               & 0.996             && 0.997\\
            \textbf{Gradient Boosting Classifier}    & \textbf{0.999}      & \textbf{0.999}    && \textbf{0.999}\\
            X-Gradient Boosting Classifier          & 0.999               & 0.998             && 0.999\\
            Light Gradient Boosting Classifier      & 0.998               & 0.998             && 0.998\\
        \bottomrule
    \end{tabular}
}
\end{table}

\begin{figure}[H]
    \begin{center}
        \includegraphics[width=1\textwidth]{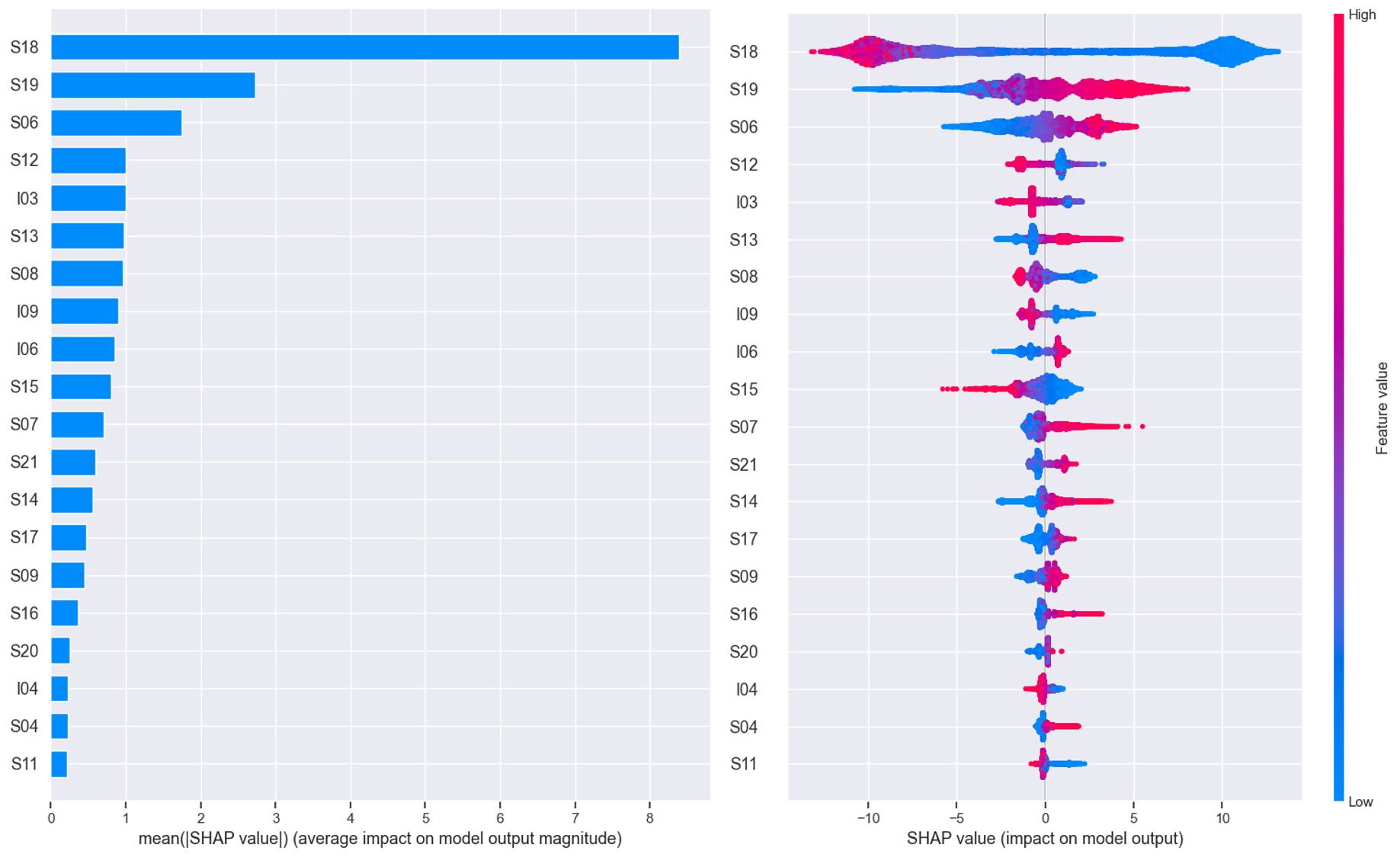}
        \caption{Plot of scenario 3 by using Gradient Boosting classifier with $F_1$-score $= 0.999$.}
        \label{img:app-3}
    \end{center}
\end{figure}

\newpage
\subsection{Scenario 4}
\label{subsec:app-scr4}

\begin{table}[htbp]
\caption{Performance of several baseline classifier used in the scenario 4.}
\label{table:f1-scr4}
\vspace*{0.1cm}
\centering
\scalebox{0.88}
{
    \begin{tabular}{lcccc}
        \toprule
            \textbf{Algorithm}                      & \textbf{Precision}  & \textbf{Recall}   && \textbf{$F_1$-score}\\
        \midrule
            K-Nearnest Neighbors Classifier         & 0.430               & 0.647             && 0.517\\
            Decision Tree Classifier                & 0.545               & 0.732             && 0.625\\
            Random Forest Classifier                & 0.572               & 0.910             && 0.702\\
            Gradient Boosting Classifier          & 0.676               & 0.907             && 0.775\\
            \textbf{X-Gradient Boosting Classifier}$^*$ & \textbf{0.723}      & \textbf{0.837}    && \textbf{0.775}\\
            Light Gradient Boosting Classifier      & 0.660              & 0.904             && 0.763\\
        \bottomrule
    \end{tabular}
}
\end{table}

\begin{figure}[H]
    \begin{center}
        \includegraphics[width=1\textwidth]{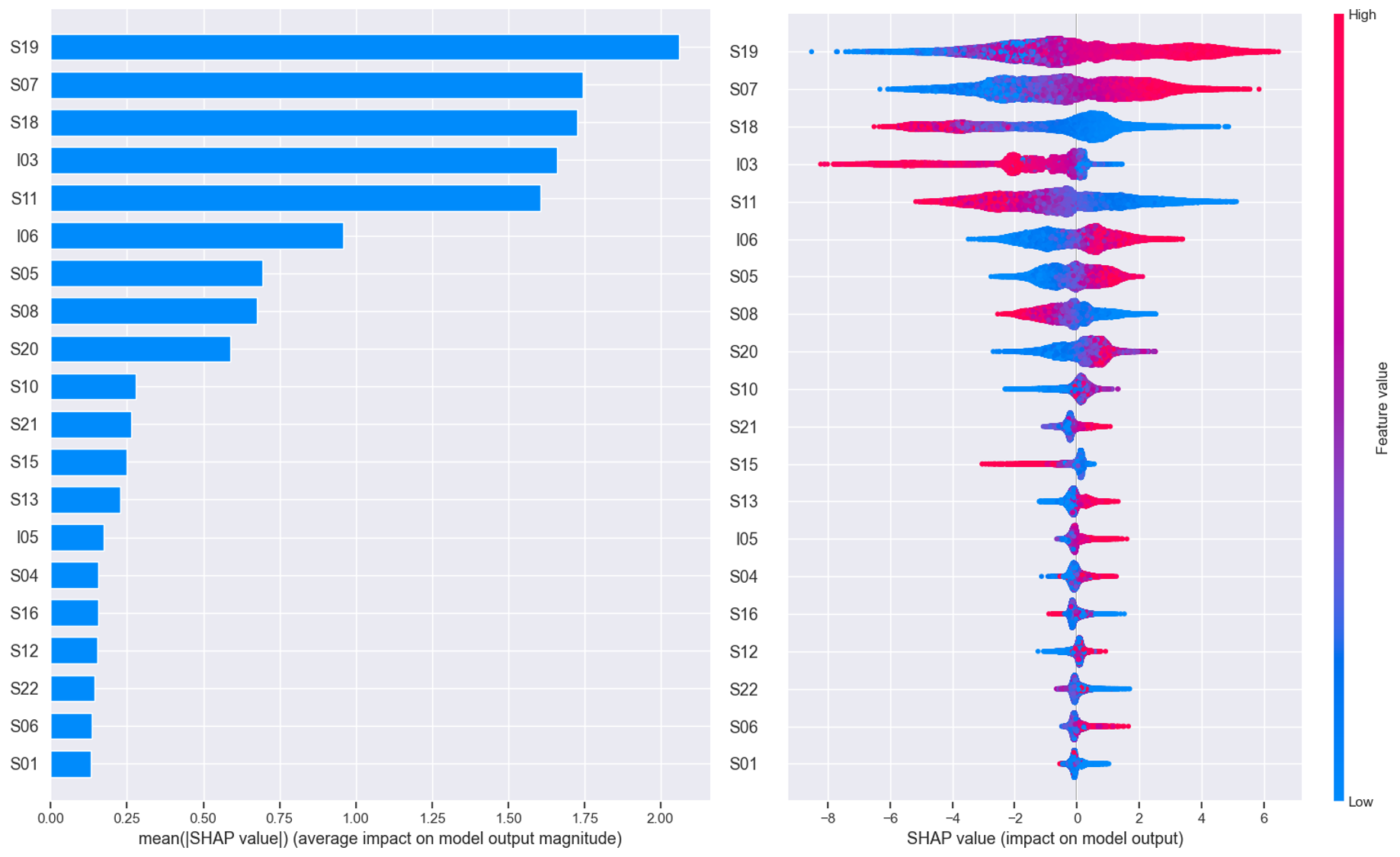}
        \caption{Plot of scenario 4 by using X-Gradient Boosting classifier with $F_1$-score $= 0.775$.}
        \label{img:app-4}
    \end{center}
\end{figure}

\newpage
\subsection{Scenario 5}
\label{subsec:app-scr5}

\begin{table}[htbp]
\caption{Performance of several baseline classifier used in the scenario 5.}
\label{table:f1-scr5}
\vspace*{0.1cm}
\centering
\scalebox{0.88}
{
    \begin{tabular}{lcccc}
        \toprule
            \textbf{Algorithm}                      & \textbf{Precision}  & \textbf{Recall}   && \textbf{$F_1$-score}\\
        \midrule
            K-Nearnest Neighbors Classifier         & 0.620               & 0.769             && 0.686\\
            Decision Tree Classifier                & 0.765               & 0.828             && 0.791\\
            Random Forest Classifier                & 0.827               & 0.892             && 0.858\\
            Gradient Boosting Classifier          & 0.863               & 0.963             && 0.910\\
            \textbf{X-Gradient Boosting Classifier} & \textbf{0.895}      & \textbf{0.951}    && \textbf{0.922}\\
            Light Gradient Boosting Classifier      & 0.856               & 0.961             && 0.906\\
        \bottomrule
    \end{tabular}
}
\end{table}

\begin{figure}[H]
    \begin{center}
        \includegraphics[width=1\textwidth]{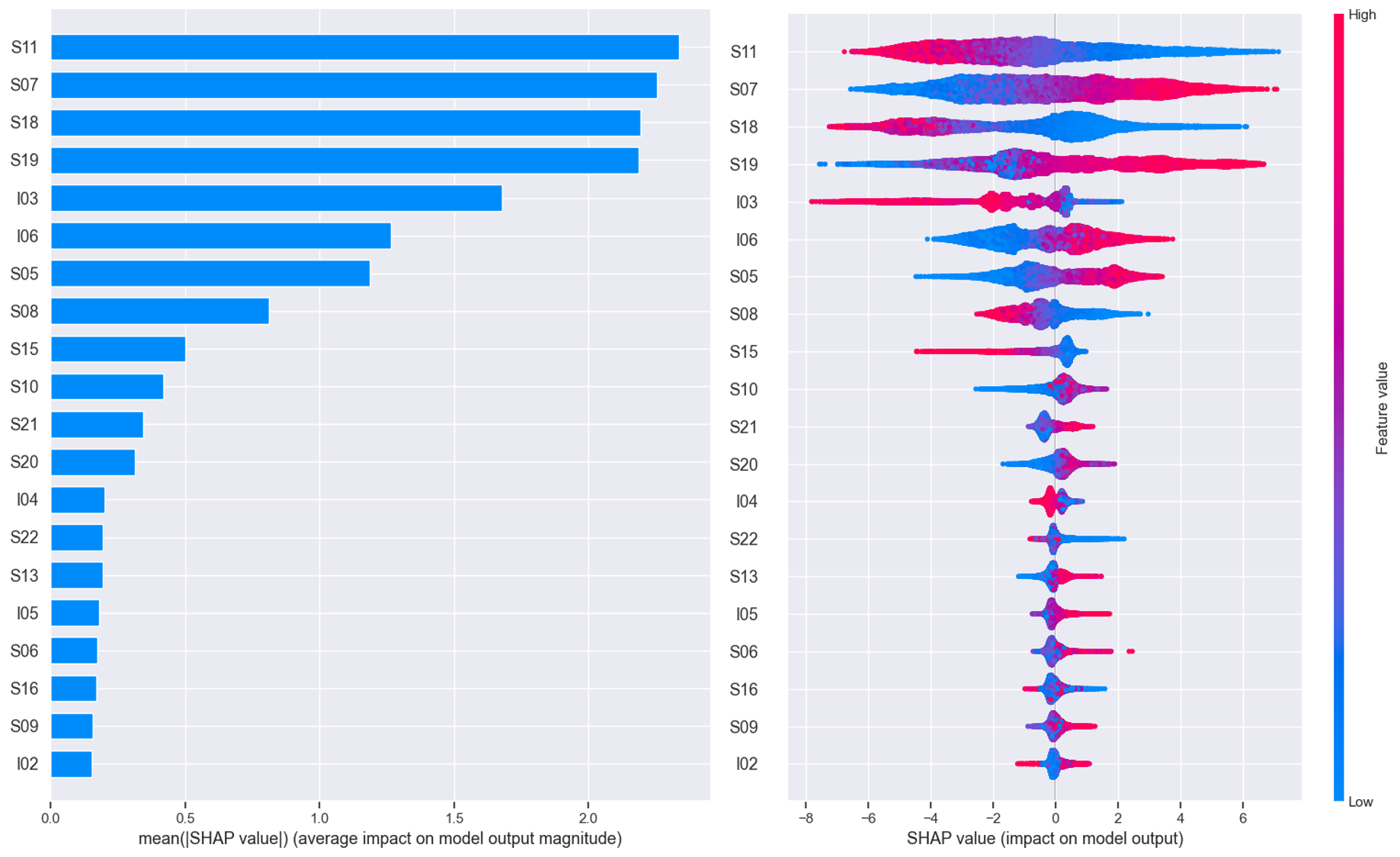}
        \caption{Plot of scenario 5 by using X-Gradient Boosting classifier with $F_1$-score $= 0.922$.}
        \label{img:app-5}
    \end{center}
\end{figure}

\newpage
\subsection{Scenario 6}
\label{subsec:app-scr6}

\begin{table}[htbp]
\caption{Performance of several baseline classifier used in the scenario 6.}
\label{table:f1-scr6}
\vspace*{0.1cm}
\centering
\scalebox{0.88}
{
    \begin{tabular}{lcccc}
        \toprule
            \textbf{Algorithm}                      & \textbf{Precision}  & \textbf{Recall}   && \textbf{$F_1$-score}\\
        \midrule
            K-Nearnest Neighbors Classifier         & 0.720               & 0.824             && 0.768\\
            Decision Tree Classifier                & 0.841               & 0.881             && 0.863\\
            Random Forest Classifier                & 0.887               & 0.962             && 0.923\\
            Gradient Boosting Classifier          & 0.924               & 0.979             && 0.951\\
            \textbf{X-Gradient Boosting Classifier} & \textbf{0.941}      & \textbf{0.978}    && \textbf{0.959}\\
            Light Gradient Boosting Classifier      & 0.924               & 0.981             && 0.952\\
        \bottomrule
    \end{tabular}
}
\end{table}

\begin{figure}[H]
    \begin{center}
        \includegraphics[width=1\textwidth]{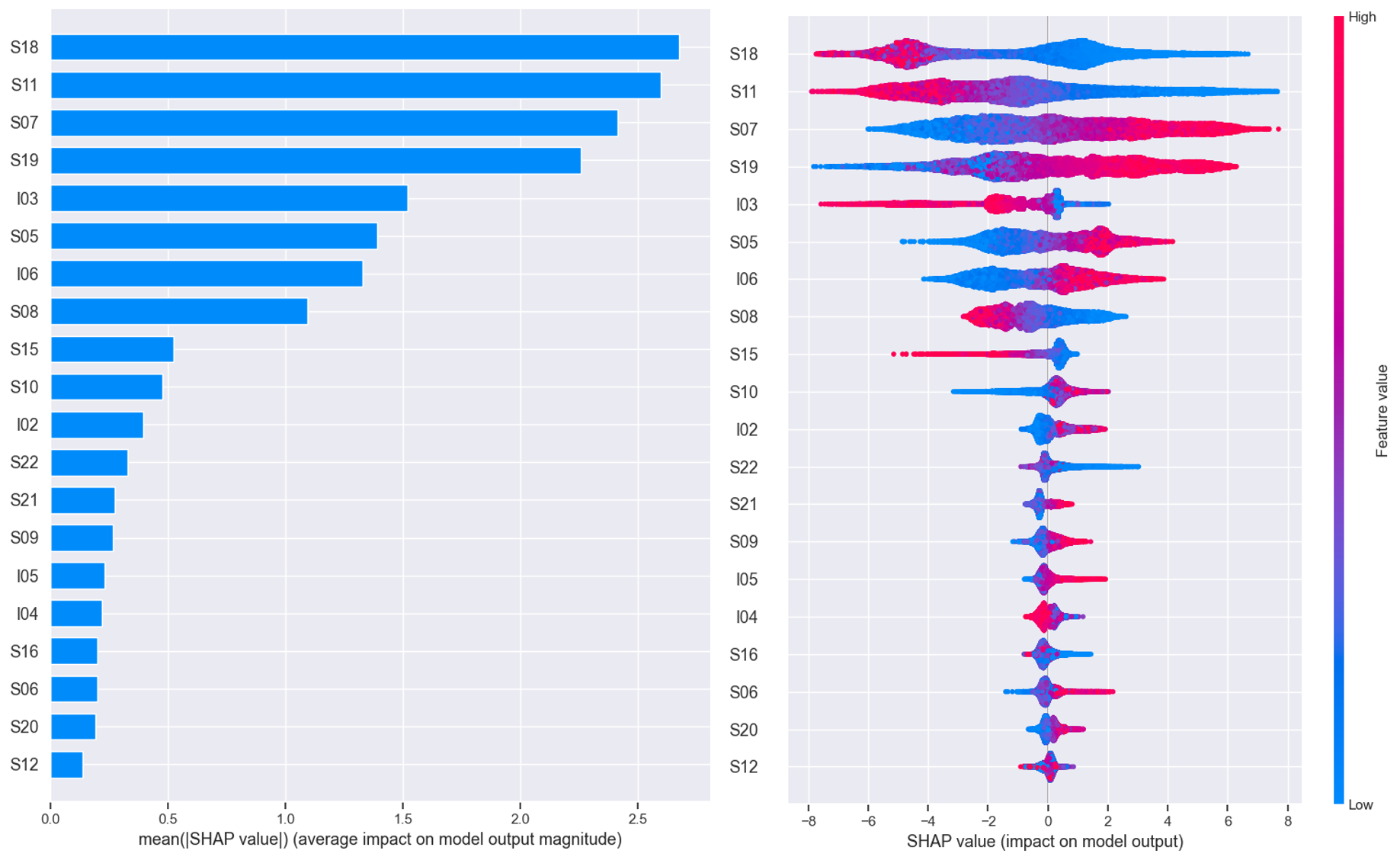}
        \caption{Plot of scenario 6 by using X-Gradient Boosting classifier with $F_1$-score $= 0.959$.}
        \label{img:app-6}
    \end{center}
\end{figure}

\newpage
\subsection{Scenario 7}
\label{subsec:app-scr7}

\begin{table}[htbp]
\caption{Performance of several baseline classifier used in the scenario 7.}
\label{table:f1-scr7}
\vspace*{0.1cm}
\centering
\scalebox{0.88}
{
    \begin{tabular}{lcccc}
        \toprule
            \textbf{Algorithm}                      & \textbf{Precision}  & \textbf{Recall}   && \textbf{$F_1$-score}\\
        \midrule
            K-Nearnest Neighbors Classifier         & 0.827               & 0.871             && 0.848\\
            Decision Tree Classifier                & 0.928               & 0.933             && 0.930\\
            Random Forest Classifier                & 0.936               & 0.991             && 0.962\\
            Gradient Boosting Classifier          & 0.966               & 0.991             && 0.979\\
            \textbf{X-Gradient Boosting Classifier} & \textbf{0.972}      & \textbf{0.994}    && \textbf{0.983}\\
            Light Gradient Boosting Classifier      & 0.967               & 0.994             && 0.980\\
        \bottomrule
    \end{tabular}
}
\end{table}

\begin{figure}[H]
    \begin{center}
        \includegraphics[width=1\textwidth]{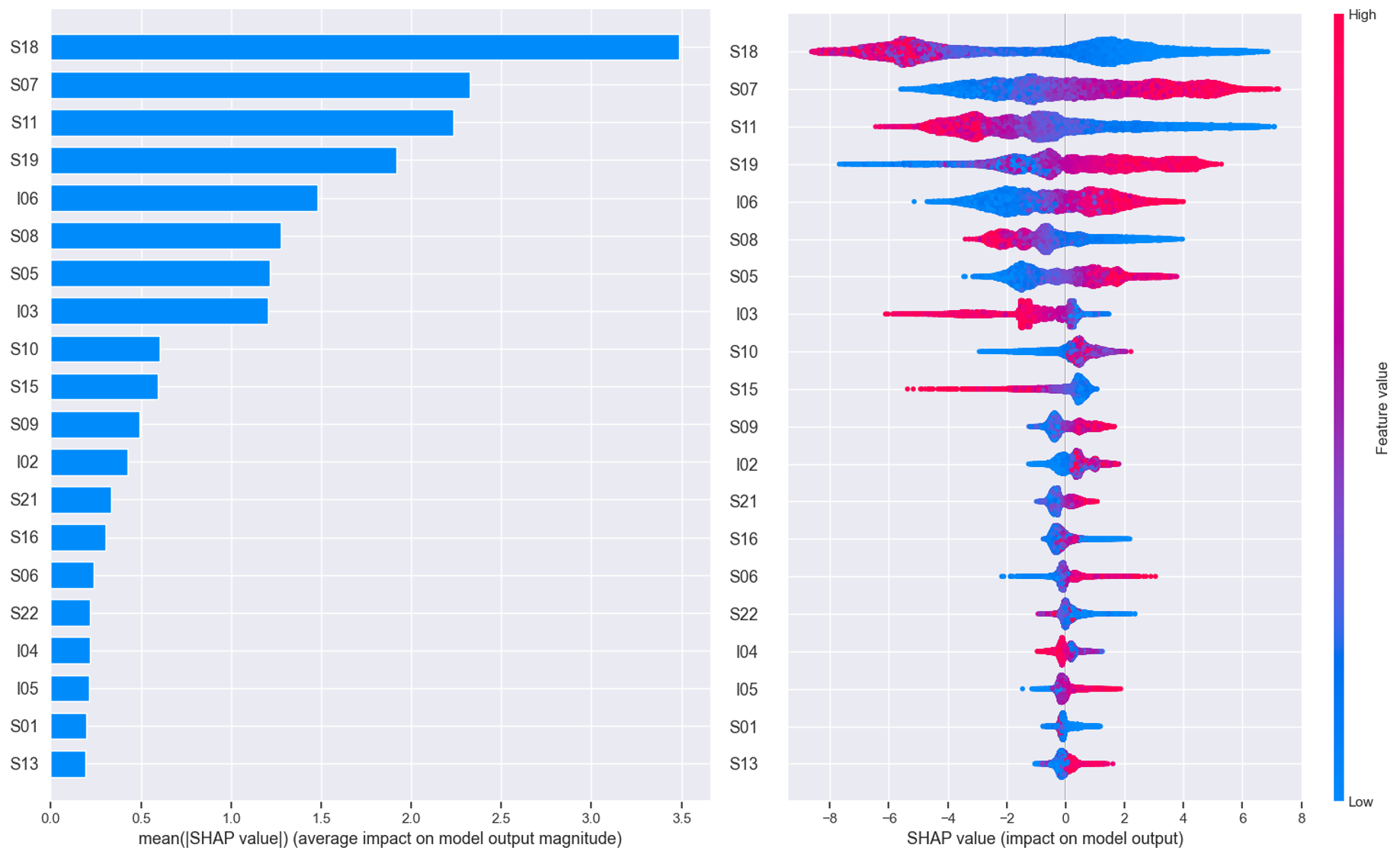}
        \caption{Plot of scenario 7 by using X-Gradient Boosting classifier with $F_1$-score $= 0.983$.}
        \label{img:app-7}
    \end{center}
\end{figure}

\newpage
\subsection{Scenario 8}
\label{subsec:app-scr8}

\begin{table}[htbp]
\caption{Performance of several baseline classifier used in the scenario 8.}
\label{table:f1-scr8}
\vspace*{0.1cm}
\centering
\scalebox{0.88}
{
    \begin{tabular}{lcccc}
        \toprule
            \textbf{Algorithm}                      & \textbf{Precision}  & \textbf{Recall}   && \textbf{$F_1$-score}\\
        \midrule
            K-Nearnest Neighbors Classifier         & 0.895               & 0.896             && 0.895\\
            Decision Tree Classifier                & 0.980               & 0.976             && 0.978\\
            Random Forest Classifier                & 0.985               & 0.998             && 0.991\\
            \textbf{Gradient Boosting Classifier}$^*$    & \textbf{0.993}      & \textbf{0.999}    && \textbf{0.996}\\
            X-Gradient Boosting Classifier          & 0.992               & 0.998             && 0.996\\
            Light Gradient Boosting Classifier      & 0.992               & 0.998             && 0.995\\
        \bottomrule
    \end{tabular}
}
\end{table}

\begin{figure}[H]
    \begin{center}
        \includegraphics[width=1\textwidth]{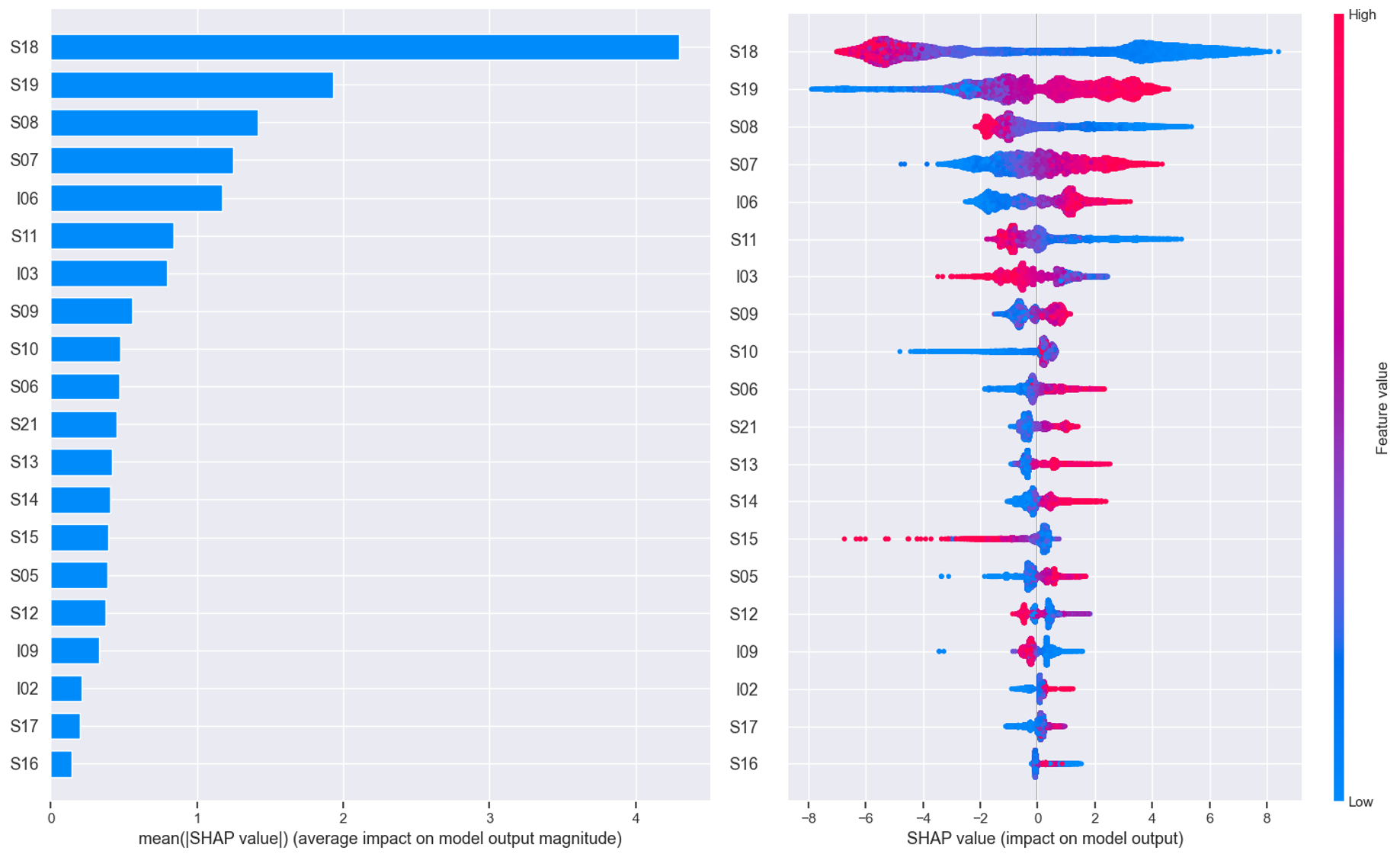}
        \caption{Plot of scenario 8 by using Gradient Boosting classifier with $F_1$-score $= 0.996$.}
        \label{img:app-8}
    \end{center}
\end{figure}

\clearpage
\printbibliography

@article{de2024improved,
title={An improved hybrid genetic search with data mining for the CVRP},
author={de Holanda Maia, Marcelo Rodrigues and Plastino, Alexandre and dos Santos Souza, U{\'e}verton},
journal={Networks},
volume={4},
number={83},
pages={692--711},
year={2024},
doi="10.1002/net.22213"
}

@article{de2022interpretable,
title={Interpretable ensembles of classifiers for uncertain data with bioinformatics applications},
author={de Holanda Maia, Marcelo Rodrigues and Plastino, Alexandre and Freitas, Alex and de Magalhaes, Joao Pedro},
journal={IEEE/ACM Transactions on Computational Biology and Bioinformatics},
volume={20},
number={3},
pages={1829--1841},
year={2022},
publisher={IEEE},
doi="10.1109/TCBB.2022.3218588"
}

@article{de2020minereduce,
title={MineReduce: An approach based on data mining for problem size reduction},
author={de Holanda Maia, Marcelo Rodrigues and Plastino, Alexandre and Penna, Puca Huachi Vaz},
journal={Computers \& Operations Research},
volume={122},
pages={104995},
year={2020},
doi="10.1016/j.cor.2020.104995"
}

@inproceedings{zern2023interventional,
title={Interventional shap values and interaction values for piecewise linear regression trees},
author={Zern, Artjom and Broelemann, Klaus and Kasneci, Gjergji},
booktitle={Proceedings of the AAAI Conference on Artificial Intelligence},
pages={11164--11173},
year={2023},
doi="10.1609/aaai.v37i9.26322"
}

@inproceedings{liu2023shapley,
title={Shapley based residual decomposition for instance analysis},
author={Liu, Tommy and Barnard, Amanda S},
booktitle = {Proceedings of the International Conference on Machine Learning},
pages={21375--21387},
year={2023},
url="https://proceedings.mlr.press/v202/liu23b.html"
}

@article{chen2023algorithms,
title={Algorithms to estimate Shapley value feature attributions},
author={Chen, Hugh and Covert, Ian C and Lundberg, Scott M and Lee, Su-In},
journal={Nature Machine Intelligence},
volume={5},
number={6},
pages={590--601},
year={2023},
publisher={Nature Publishing Group UK London},
doi="10.1038/s42256-023-00657-x"
}

@article{chen2022explaining,
title={Explaining a series of models by propagating Shapley values},
author={Chen, Hugh and Lundberg, Scott M and Lee, Su-In},
journal={Nature communications},
volume={13},
number={1},
pages={4512},
year={2022},
publisher={Nature Publishing Group UK London},
doi = "10.1038/s41467-022-31384-3"
}

@article{sokolova2009systematic,
title={A systematic analysis of performance measures for classification tasks},
author={Sokolova, Marina and Lapalme, Guy},
journal={Information processing \& management},
volume={45},
number={4},
pages={427--437},
year={2009},
doi="10.1016/j.ipm.2009.03.002"
}

@inproceedings{davis2006relationship,
title={The relationship between Precision-Recall and ROC curves},
author={Davis, Jesse and Goadrich, Mark},
booktitle = {Proceedings of the International Conference on Machine Learning},
pages={233--240},
year={2006},
doi="10.1145/1143844.1143874"
}

@article{narasimhan2014statistical,
title={On the statistical consistency of plug-in classifiers for non-decomposable performance measures},
author={Narasimhan, Harikrishna and Vaish, Rohit and Agarwal, Shivani},
journal={Advances in neural information processing systems},
url = {https://proceedings.neurips.cc/paper_files/paper/2014/file/32b30a250abd6331e03a2a1f16466346-Paper.pdf},
volume = {27},
year = {2014}
}

@article{ke2017lightgbm,
title={Lightgbm: A highly efficient gradient boosting decision tree},
author={Ke, Guolin and Meng, Qi and Finley, Thomas and Wang, Taifeng and Chen, Wei and Ma, Weidong and Ye, Qiwei and Liu, Tie-Yan},
journal={Advances in neural information processing systems},
volume={30},
year={2017},
url="https://dl.acm.org/doi/10.5555/3294996.3295074"
}

@inproceedings{chen2016xgboost,
title={Xgboost: A scalable tree boosting system},
author={Chen, Tianqi and Guestrin, Carlos},
booktitle = {Proceedings of the {ACM SIGKDD} Conference on Knowledge Discovery and Data Mining},
pages={785--794},
year={2016},
doi="10.1145/2939672.2939785"
}

@article{friedman2001greedy,
title={Greedy function approximation: a gradient boosting machine},
author={Friedman, Jerome H},
journal={Annals of statistics},
pages={1189--1232},
year={2001},
publisher={JSTOR},
URL = {http://www.jstor.org/stable/2699986}
}

@article{breiman2001random,
title={Random forests},
author={Breiman, Leo},
journal={Machine learning},
volume={45},
pages={5--32},
year={2001},
publisher={Springer},
doi="10.1023/A:1010933404324"
}

@article{quinlan1986induction,
title={Induction of decision trees},
author={Quinlan, J. Ross},
journal={Machine learning},
volume={1},
pages={81--106},
year={1986},
publisher={Springer},
doi={https://doi.org/10.1007/BF00116251}
}

@book{fix1985discriminatory,
title={Discriminatory analysis: nonparametric discrimination, consistency properties},
author={Fix, Evelyn},
volume={1},
year={1985},
publisher={USAF school of Aviation Medicine}
}

@article{gillett1974heuristic,
title={A heuristic algorithm for the vehicle-dispatch problem},
author={Gillett, Billy E and Miller, Leland R},
journal={Operations research},
volume={22},
number={2},
pages={340--349},
year={1974},
publisher={INFORMS},
doi="10.1287/opre.22.2.340"
}

@inproceedings{cavalcanti2023learning,
title={Learning to select initialisation heuristic for vehicle routing problems},
author={Cavalcanti Costa, Joao Guilherme and Mei, Yi and Zhang, Mengjie},
booktitle={Proceedings of the Genetic and Evolutionary Computation Conference},
pages={266--274},
year={2023},
doi="10.1145/3583131.3590397"
}

@article{bengio2021machine,
title={Machine learning for combinatorial optimization: a methodological tour d’horizon},
author={Bengio, Yoshua and Lodi, Andrea and Prouvost, Antoine},
journal={European Journal of Operational Research},
volume={290},
number={2},
pages={405--421},
year={2021},
doi="10.1016/j.ejor.2020.07.063"
}

@article{li2021learning,
title={Learning to delegate for large-scale vehicle routing},
author={Li, Sirui and Yan, Zhongxia and Wu, Cathy},
journal={Advances in Neural Information Processing Systems},
volume={34},
pages={26198--26211},
year={2021},
url="https://openreview.net/forum?id=rm0I5y2zkG8"
}

@article{lucas2019comment,
title={A comment on “what makes a {VRP} solution good? The generation of problem-specific knowledge for heuristics”},
author={Lucas, Flavien and Billot, Romain and Sevaux, Marc},
journal={Computers \& Operations Research},
volume={110},
pages={130--134},
year={2019},
doi="10.1016/j.cor.2019.05.025"
}

@article{arnold2019knowledge,
title={Knowledge-guided local search for the vehicle routing problem},
author={Arnold, Florian and S{\"o}rensen, Kenneth},
journal={Computers \& Operations Research},
volume={105},
pages={32--46},
year={2019},
doi="10.1016/j.cor.2019.01.002"
}

@article{clarke1964scheduling,
title={Scheduling of vehicles from a central depot to a number of delivery points},
author={Clarke, Geoff and Wright, John W},
journal={Operations research},
volume={12},
number={4},
pages={568--581},
year={1964},
publisher={Informs},
doi="10.1007/978-3-642-27922-5_18"
}

@article{morabit2023learning,
title={Learning to repeatedly solve routing problems},
author={Morabit, Mouad and Desaulniers, Guy and Lodi, Andrea},
journal={Networks},
year={2023},
doi="10.1002/net.22200"
}

@article{laporte2009fifty,
title={Fifty years of vehicle routing},
author={Laporte, Gilbert},
journal={Transportation science},
volume={43},
number={4},
pages={408--416},
year={2009},
publisher={INFORMS},
doi="10.1287/trsc.1090.0301"
}

@article{lundberg2020local,
title={From local explanations to global understanding with explainable {AI} for trees},
author={Lundberg, Scott M and Erion, Gabriel and Chen, Hugh and DeGrave, Alex and Prutkin, Jordan M and Nair, Bala and Katz, Ronit and Himmelfarb, Jonathan and Bansal, Nisha and Lee, Su-In},
journal={Nature machine intelligence},
volume={2},
number={1},
pages={56--67},
year={2020},
publisher={Nature Publishing Group UK London},
doi="10.1038/s42256-019-0138-9"
}

@inproceedings{lucas2020reducing,
title={Reducing space search in combinatorial optimization using machine learning tools},
author={Lucas, Flavien and Billot, Romain and Sevaux, Marc and S{\"o}rensen, Kenneth},
booktitle={Learning and Intelligent Optimization: 14th International Conference, LION 14, Athens, Greece, May 24--28, 2020, Revised Selected Papers 14},
pages={143--150},
year={2020},
organization={Springer},
doi="10.1007/978-3-030-53552-0_15"
}

@article{queiroga202110,
title={10,000 optimal {CVRP} solutions for testing machine learning based heuristics},
author={Queiroga, Eduardo and Sadykov, Ruslan and Uchoa, Eduardo and Vidal, Thibaut},
journal={AAAI-22 Workshop on Machine Learning for Operations Research (ML4OR)},
year={2021},
url="https://openreview.net/forum?id=yHiMXKN6nTl"
}

@inproceedings{
ma2023learning,
title={Learning to Search Feasible and Infeasible Regions of Routing Problems with Flexible Neural k-Opt},
author={Yining Ma and Zhiguang Cao and Yeow Meng Chee},
booktitle={Advances in Neural Information Processing Systems},
year={2023},
url={https://openreview.net/forum?id=q1JukwH2yP}
}

@article{xin2021neurolkh,
title={Neuro{LKH}: Combining deep learning model with lin-kernighan-helsgaun heuristic for solving the traveling salesman problem},
author={Xin, Liang and Song, Wen and Cao, Zhiguang and Zhang, Jie},
journal={Advances in Neural Information Processing Systems},
volume={34},
pages={7472--7483},
year={2021},
url="https://openreview.net/forum?id=VKVShLsAuZ"
}

@article{morabit2021machine,
title={Machine-learning--based column selection for column generation},
author={Morabit, Mouad and Desaulniers, Guy and Lodi, Andrea},
journal={Transportation Science},
volume={55},
number={4},
pages={815--831},
year={2021},
publisher={INFORMS},
doi="10.1287/trsc.2021.1045"
}

@article{arnold2019makes,
title={What makes a {VRP} solution good? The generation of problem-specific knowledge for heuristics},
author={Arnold, Florian and S{\"o}rensen, Kenneth},
journal={Computers \& Operations Research},
volume={106},
pages={280--288},
year={2019},
doi="10.1016/j.cor.2018.02.007"
}

@Inbook{Prodhon2016,
author={Prodhon, Caroline and Prins, Christian},
title="Metaheuristics for Vehicle Routing Problems",
bookTitle="Metaheuristics",
year="2016",
publisher="Springer International Publishing",
address="Cham",
pages="407--437",
doi="10.1007/978-3-319-45403-0_15"
}

@article{soto2017multiple,
title={Multiple neighborhood search, tabu search and ejection chains for the multi-depot open vehicle routing problem},
author={Soto, Mar{\'\i}a and Sevaux, Marc and Rossi, Andr{\'e} and Reinholz, Andreas},
journal={Computers \& Industrial Engineering},
volume={107},
pages={211--222},
year={2017},
doi="10.1016/j.cie.2017.03.022"
}

@article{lundberg2017unified,
title={A unified approach to interpreting model predictions},
author={Lundberg, Scott M and Lee, Su-In},
journal={Advances in neural information processing systems},
volume={30},
year={2017},
url="https://proceedings.neurips.cc/paper_files/paper/2017/file/8a20a8621978632d76c43dfd28b67767-Paper.pdf"
}

@article{baptista2022relation,
title={Relation between prognostics predictor evaluation metrics and local interpretability SHAP values},
author={Baptista, Marcia L and Goebel, Kai and Henriques, Elsa MP},
journal={Artificial Intelligence},
volume={306},
pages={103667},
year={2022},
doi="10.1016/j.artint.2022.103667"
}

@article{ARNOLD201932,
title = {Efficiently solving very large-scale routing problems},
journal = {Computers \& Operations Research},
volume = {107},
pages = {32-42},
year = {2019},
issn = {0305-0548},
doi = "10.1016/j.cor.2019.03.006",
author = {Florian Arnold and Michel Gendreau and Kenneth Sörensen}
}

@article{guidotti2018survey,
title={A survey of methods for explaining black box models},
author={Guidotti, Riccardo and Monreale, Anna and Ruggieri, Salvatore and Turini, Franco and Giannotti, Fosca and Pedreschi, Dino},
journal={ACM computing surveys (CSUR)},
doi="10.1145/3236009",
volume={51},
number={5},
pages={1--42},
year={2018},
publisher={ACM New York, NY, USA}
}

@inproceedings{santana2023neural,
title={Neural Networks for Local Search and Crossover in Vehicle Routing: A Possible Overkill?},
author={Santana, {\'I}talo and Lodi, Andrea and Vidal, Thibaut},
booktitle={International Conference on Integration of Constraint Programming, Artificial Intelligence, and Operations Research},
pages={184--199},
year={2023},
organization={Springer},
doi="10.1007/978-3-031-33271-5_13"
}

@article{groer2010library,
title={A library of local search heuristics for the vehicle routing problem},
author={Gro{\"e}r, Chris and Golden, Bruce and Wasil, Edward},
journal={Mathematical Programming Computation},
volume={2},
pages={79--101},
year={2010},
publisher={Springer},
doi="10.1007/s12532-010-0013-5"
}

@article{mesa2022machine,
title = {Machine-learning component for multi-start metaheuristics to solve the capacitated vehicle routing problem},
journal = {Applied Soft Computing},
volume = {173},
pages = {112916},
year = {2025},
issn = {1568-4946},
doi = {https://doi.org/10.1016/j.asoc.2025.112916},
author = {Juan Pablo Mesa and Alejandro Montoya and Raul Ramos-Pollan and Mauricio Toro}
}

@article{herdianto:hal-04310676,
  title={Metaheuristic Enhanced with Feature-Based Guidance and Diversity Management for Solving the Capacitated Vehicle Routing Problem}, 
  author={Bachtiar Herdianto and Romain Billot and Flavien Lucas and Marc Sevaux},
  year={2024},
  url={https://arxiv.org/abs/2407.20777}, 
  journal={arXiv preprint arXiv:2407.20777}
}

\end{document}